\let\mypdfximage\pdfximage
\def\pdfximage{\immediate\mypdfximage}
\pdfoutput=1

\documentclass[times,twocolumn,review]{elsarticle}
\usepackage{medima}

\usepackage{hyperref}
\usepackage[export]{adjustbox}
\usepackage[cmex10]{amsmath}
\usepackage{amssymb, amsfonts}
\usepackage{extarrows}
\usepackage{array,multirow}
\usepackage{color, colortbl}
\usepackage{makeidx}
\usepackage{rotating}
\usepackage{graphicx}
\usepackage{mathtools}
\usepackage{bbm}
\usepackage{tikz}
\usepackage[multiple]{footmisc}
\tikzstyle{vertex}=[auto=left,circle,fill=black!25,minimum size=20pt,inner sep=0pt]
\usepackage{enotez}

\definecolor{col1}{rgb}{0.988,0.839,0.571}
\definecolor{col2}{rgb}{0.886,0.886,0.555}
\definecolor{col3}{rgb}{0.582,0.661,0.976}
\definecolor{white}{rgb}{1,1,1}

\graphicspath{{./Figures/},{./Figures/OasisCor/},{./Figures/OasisRegression/},{./Figures/Ellipse/}}
\DeclareGraphicsExtensions{.pdf,.png}

\newcommand{\cor}[0]{\mathtt{cor}}

\newcommand{\cost}[0]{\mathtt{c}}
\newcommand{\acost}[0]{\mathtt{c_a}}

\newcommand{\coupling}[0]{\pi}
\newcommand{\allocation}[0]{\tau}

\def\argmin{\mathop{\rm argmin}}

\newcommand{\Xsp}{{\mathbf{X}}}
\newcommand{\Ysp}{{\mathbf{Y}}}

\newcommand{\AgeH}{Age\textsubscript{H}}
\newcommand{\AgeD}{Age\textsubscript{D}}
\newcommand{\CDRvm}{CDR\textsubscript{v-mild} }
\newcommand{\CDRm}{CDR\textsubscript{mild} }
\newcommand{\TBMA}{TBM\textsubscript{A}}
\newcommand{\TBMT}{TBM\textsubscript{T}}
\newcommand{\VBMA}{VBM\textsubscript{A}}
\newcommand{\VBMT}{VBM\textsubscript{T}}

\usepackage[normalem]{ulem}

\definecolor{hlAddColor}{rgb}{0.65, 1.0, 0.65}
\definecolor{hlDelColor}{rgb}{1.0, 0.65, 0.65}
\definecolor{txtAddColor}{rgb}{0.0, 0.5, 0.0}
\definecolor{txtDelColor}{rgb}{0.5, 0.0, 0.0}
\definecolor{hlRevColor}{rgb}{1.0, 1.0, 0.4}
\definecolor{txtRevColor}{rgb}{0.76, 0.13, 0.28}
\newcommand\hlRev{\bgroup\markoverwith
  {\color{hlRevColor}{\rule[-.5ex]{2pt}{2.5ex}}}\ULon}
\newcommand\hlAdd{\bgroup\markoverwith
  {\color{hlAddColor}{\rule[-.5ex]{2pt}{2.5ex}}}\ULon}

\newcommand{\rev}[1]{{\color{txtRevColor}#1}}
\DeclarePairedDelimiter\ceiling{\lceil}{\rceil}
\DeclarePairedDelimiter\floor{\lfloor}{\rfloor}

\renewcommand{\rev}[1]{{#1}} % Disable coloring of revised text

\begin{document}

\verso{Samuel Gerber \textit{et~al.}}
\begin{frontmatter}

\title{Optimal Transport Features for Morphometric Population Analysis}
\author[1]{Samuel \snm{Gerber}}
\author[2]{Marc  \snm{Niethammer}}
\author[1]{Ebrahim  \snm{Ebrahim}\corref{cor1}}
\cortext[cor1]{Corresponding author: ebrahim.ebrahim@kitware.com}
\author[2]{Joseph  \snm{Piven}}
\author[4]{Stephen R.  \snm{Dager}}
\author[2]{Martin  \snm{Styner}}
\author[1]{Stephen  \snm{Aylward}}
\author[1]{Andinet  \snm{Enquobahrie}}

\address[1]{Kitware Inc., NC, USA}
\address[2]{University of North Carolina, Chapel Hill, NC, USA}
%\address[3]{New York University, New York City, NY, USA}
\address[4]{University of Washington, Seattle, WA, USA}
%\address[5]{Washington University in St. Louis, MO, USA}

%\author{Samuel Gerber, Marc Niethammer, Guido Gerig, %
%        Joseph Piven, Stephen R. Dager,  Martin Styner, %
%        Stephen Aylward, Andinet Enquobahrie, Beau M. Ances%
%\thanks{S. Gerber, S. Aylward and A. Enquobahrie are with Kitware Inc., NC, USA; %
%        M. Niethammer, J. Piven and M. Styner are with %
%        the University of North Carolina, Chapel Hill, NC, USA; %
%        G. Gerig is with the New York University, New York City, NY, USA; %
%        S. Dager is with the University of Washington, Seattle, WA, USA; %
%        B. Ances is with the Washington University in St. Louis, MO, USA %
%        }
%\thanks{This work was funded in part, by NIH grants R01EB021391, R41MH118845-01,
%R01HD055741, U54HD079124, R42NS086295, R44NS081792, R44CA165621, and
%R01EB021396 and by NSF grant ECCS-1711776.}
%}

%\markboth{Transactions on Medical Imaging,~Vol.~?, No.~?, ???~2019}%
%{Gerber \MakeLowercase{\textit{et al.}}: Optimal Transport Morphometry}

%\maketitle

\begin{abstract}
 Brain pathologies often manifest as partial or complete loss of tissue. The
  goal of many neuroimaging studies is to capture the location and amount of
  tissue changes with respect to a clinical variable of interest,
  such as disease progression. Morphometric analysis
  approaches capture local differences in the distribution of tissue or other
  quantities of interest in relation to a clinical variable.  We propose to
  augment morphometric analysis with an additional feature extraction step
  based on unbalanced optimal transport. The optimal transport feature
  extraction step increases statistical power for pathologies that cause
  spatially dispersed tissue loss, minimizes sensitivity to shifts due to
  spatial misalignment or differences in brain topology, and separates changes
  due to volume differences from changes due to tissue location. We demonstrate
  the proposed optimal transport feature extraction step in the context of a volumetric
  morphometric analysis of the OASIS-1 study for Alzheimer's disease.  The
  results demonstrate that the proposed approach can identify tissue changes
  and differences that are not otherwise measurable.
\end{abstract}

%\begin{IEEEkeywords}
%\end{IEEEkeywords}

%\IEEEpeerreviewmaketitle

\end{frontmatter}

%\linenumbers
\section{Introduction}
Neurological diseases and disorders produce subtle and varied changes in
brain anatomy that can be diffuse in nature and affect tissue volume as well
as the relative position and shape of brain anatomy. Detecting and quantifying
these changes are the primary goals of image-based morphometric population
analysis studies.  Popular methods include cortical-surface-based
analysis~\citep{dale1999cortical} or volumetric methods such as voxel-based
morphometry (VBM)~\citep{ashburner2000voxel} or tensor-based morphometry
(TBM)~\citep{ashburner2007fast,hua2008tensor} and its variants.  Morphometric
methods have been applied to a wide variety of neuroimaging studies ranging
from Alzheimer's disease~\citep{hua2008tensor} to
schizophrenia~\citep{scarpazza2016voxel}.

The standard morphometric analysis pipeline for both surface and volumetric
analysis is based on three main steps: (1) spatial alignment, (2) extraction of
the quantity of interest, and (3) statistical analysis on the spatial domain.
While the specifics vary greatly for different methods, the main steps remain
the same and the statistical analysis is based on the analysis of quantities
defined at discrete locations on a spatial domain: per voxel in volumetric
approaches and per vertex for surface based approaches.
We propose adding optimal transport features (OTF) as an additional feature
extraction step that precedes the statistical analysis.
This can increase statistical power for the discovery of spatially diffuse effects,
it can reduce sensitivity to spatial misalignments, and it can separate changes
in quantity from changes due to shifts in tissue location.

\begin{figure}[tbh]
\footnotesize
\setlength{\tabcolsep}{0.5pt}
\renewcommand{\arraystretch}{0.4}
\centering
\begin{tabular}{ccc @{\hspace{1mm}} c @{\hspace{1mm}} ccc}
  \multicolumn{3}{c}{TBM} & \hspace{1mm}& \multicolumn{3}{c}{TBM with OTF} \\[1mm]
  \includegraphics[height=0.15\linewidth]{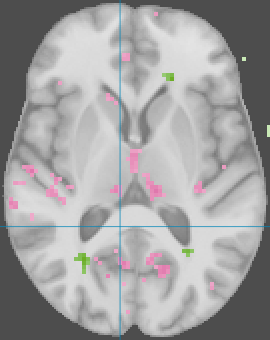} &
  \includegraphics[height=0.15\linewidth]{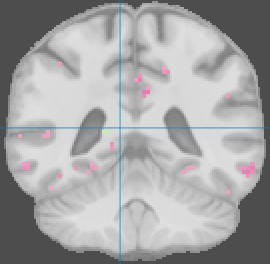} &
  \includegraphics[height=0.15\linewidth]{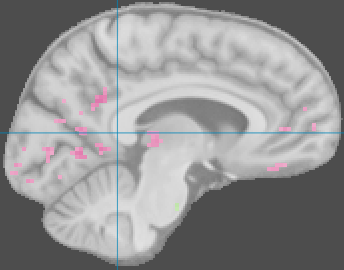} & &
  \includegraphics[height=0.15\linewidth]{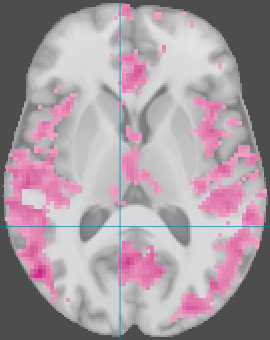} &
  \includegraphics[height=0.15\linewidth]{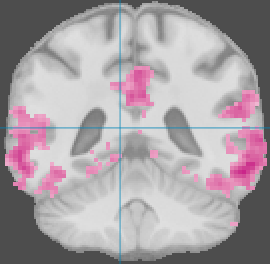} &
  \includegraphics[height=0.15\linewidth]{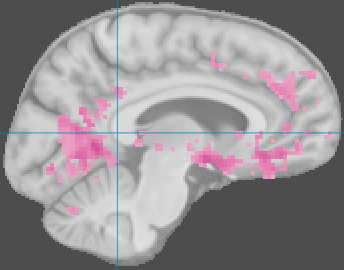} \\[2mm]
  \multicolumn{7}{c}{-0.65 \includegraphics[width=60mm, height=2.5mm, angle=0]{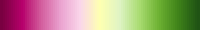} 0.65}
\end{tabular}
\caption{\label{fig:illustration}
Tensor-based morphometric analysis of gray matter tissue \rev{without and with the}
optimal transport feature extraction step. The \rev{pink} color shows
\rev{inverse} correlation strength \rev{of gray matter} between
\rev{patients with mild dementia compared to normal aging controls} for voxels with
Bonferroni corrected $p < 0.05$.  The optimal transport feature extraction
improves the statistical power and \rev{shows regions}
of gray matter tissue loss
associated with mild dementia \rev{that are not identified} with a standard TBM analysis (more
details of the analysis are in Section~\ref{sec:results}).
}
\end{figure}

The feature extraction step is based on unbalanced optimal transport and was first
introduced in the context of VBM  as unbalanced optimal transport morphometry
(UTM)~\citep{gerber2018exploratory}. We emphasize that this methodology can be used
in settings other than the VBM analysis pipelines (e.g.  surface-based) by
adding the OTF step before the statistical analysis.
This work uses volumetric-based approaches, in particular voxel- and tensor-based
morphometric (VBM, TBM), to demonstrate the properties of OTF. Voxel-based
analysis compares tissue amounts at each voxel and captures spatially localized
changes in brain anatomy.  The distribution at a particular voxel is not only
dependent on the amount of tissue but also on anatomical shape, as the gyri and sulci typically do not perfectly align
after spatial normalization
due to topological differences.
Even with a hypothetical perfect alignment, tissue
loss can be diffuse and spread out over a large region, affecting different
locations within a region per individual.  The local analysis in VBM/TBM has
difficulty in detecting such regionally or diffusely occurring individual
changes. Figure~\ref{fig:illustration} illustrates how OTF improve sensitivity
to diffuse tissue loss on a TBM  analysis of gray matter tissue volume
in correlation with clinical dementia ratings.

Optimal transport, as the name implies, solves the problem of transporting mass
from a source probability measure $\mu$ to a target probability measure $\nu$,
such that the cost of moving mass from the source $\mu$ to the target $\nu$ is
minimized.  Unbalanced optimal
transport~\citep{guittet2002extended,benamou2003numerical} extends optimal
transport to measures that do not need to have equal mass by adding a mechanism
to add (or remove) mass to the optimization problem. The solution of the
unbalanced optimal transport problem yields a transport plan, or coupling, that
measures local mass allocation as well as transport cost, i.e., the movement of
mass between source and target locations. These two measures explicitly
separate changes in amount of tissue from differences in relative location of
tissue due to variation in position and shape of individual anatomy. This
results in two main improvements over the traditional voxel-based analysis
approaches. First, the decoupling of tissue allocation and tissue movement
makes OTF less sensitive to shape variations and shifts in location of anatomy.
Second, the explicit modeling of mass allocation improves statistical power
when tissue loss (or gain) is spatially dispersed.

This paper extends the OTF approach~\citep{gerber2018exploratory} by using a
template (Section~\ref{sec:template}) and adding localized mass-balancing
(Section~\ref{sec:unbalanced}). The use of a template avoids the computation of
pairwise transport maps between all images and requires only transport maps
from each image to the template image. The original OTF formulation enforces a
global mass-balancing. This can mask regional effects of smaller magnitude if
there are large variations in total mass in the population. We propose a
mechanism to enforce mass-balancing at intermediate scales by adding a cost
to allocating or removing mass.  This localized mass-balancing approach
provides a continuum between
the VBM/TBM approach and the global mass
equalization approach in the originally proposed OTF.

We provide a statistical analysis and evaluate the method using real-world data.
The statistical analysis quantifies the increase in correlation strength for
diffuse tissue loss (Section~\ref{sec:analysis}). A key finding of the
statistical analysis is that OTF require a smaller sample size than VBM/TBM to
detect differences and yields non-negligible correlations for even very small
diffuse tissue loss differences between populations. We demonstrate the OTF analysis
approach in the context of a VBM and TBM analysis on the OASIS-1 Dementia
study data (Section~\ref{sec:results}). The OTF analysis of real world data
supports the theoretical results from Section~\ref{sec:analysis} and shows
stronger correlations for diffuse pathology. The OTF results based on real-world
data also demonstrate the capability to separate changes due to shape from
tissue loss effects.

\section{Related Work}

\rev{VBM~\citep{ashburner2000voxel} uses a smoothed tissue mask to correlate each voxel with a clinical variable of interest.
The first step of VBM is spatial alignment, but the alignment needs to be done at a global scale in order to preserve some of the local variation in tissue masks~\citep[Chapter~7]{frackowiak2004human}.
TBM~\citep{ashburner2007fast} modifies VBM by using the Jacobian determinant of
a deformable spatial alignment, rather than using a smoothed tissue mask, at each voxel to
correlate with a clinical variable of interest. The Jacobian determinant
provides a measure of local volume changes in the spatial
alignment, and this allows TBM to provide better results given better local spatial alignment.
TBM mitigates some of the difficulties
associated with choosing a spatial alignment in VBM.
However, even with state of the art deformable registration,
perfect alignment is typically not achieved, and the Jacobian
determinant depends strongly on the amount of regularization of the deformable
registration.}

Deformation-based morphometry (DBM)~\citep{ashburner1998identifying} addresses
the issue of detecting shape variations by comparing the parameters of
non-linear deformations to a template. DBM typically requires segmentation of
the anatomy of interest and the DBM results, typically shown as modes of
variation of the anatomy, \rev{are} more difficult to visualize and interpret.

The OTF analysis combines strengths of \rev{VBM/TBM} and DBM methods. While the results are based on a
voxel-wise analysis and are easily interpretable, the quantities compared stem
from a global optimization problem that can detect global and regional shape
variations and tissue loss effects.
On the other hand, VBM, TBM, and DBM are driven by local image gradients,
and voxel-wise measures ultimately lead to very similar results.
The OTF method solves a global optimization
problem based on the distribution of mass, and the allocation of mass is not
driven solely by image gradients. With OTF, allocation of mass can be diffuse over
a large region without the need for a smooth spatial normalization to
distribute the image gradient driven warp over a larger region. TBM and DBM are
not able to capture diffuse tissue loss consisting of small deteriorations
at multiple locations if the tissue loss does not affect anatomical boundaries.
VBM detects such loss but requires more data samples compared to OTF, as we
show in Section~\ref{sec:analysis}.

VBM is known to be sensitive to the particulars of the spatial
normalization~\citep{bookstein2001voxel,davatzikos2004voxel}. OTF still depend
on a spatial normalization to bring the participants into a common coordinate
system such that distances between voxels across participants are meaningful.
However, optimal transport is insensitive to small miss-alignments in the
spatial normalization step; small shifts in mass locations incur only small
transport costs.

\rev{There are several different formulations of optimal transport theory, with much recent interest in extending the formulations and algorithms to the unbalanced case.
The fluid dynamic formulation of optimal transport, introduced by~\citep{benamou:nm2000}, was extended to the unbalanced case by~\citep{Liero_2017,Chizat_2018,Gangbo_2019,Lee_2021}.
When the measures are finitely supported, the optimal transport problem
becomes equivalent to an assignment problem~\citep{Merigot_2021},
and unbalanced transport similarly becomes unbalanced assignment;
The unbalanced assignment problem has its own history of algorithm development; a summary can be found in~\citep{Prakash_2022}.
In this work, we present unbalanced transport as a linear program, 
and we frame this in our software as a minimum cost flow problem that we solve using the Network Simplex Method~\citep{Ahuja:1993:NFT:137406}.}
Recent advances in the computation of optimal transport
plans~\citep{cuturi2013sinkhorn,gerber2017multiscale} paved the way for a flurry
of applications in machine learning and spurred interest in applications to
medical image analysis. Gramfort et. al.~\citep{gramfort2015fast} used optimal
transport for improved averaging of neuroimaging data, Feydy et
al.~\citep{feydy2017optimal} used unbalanced optimal transport as a similarity
measure for diffeomorphic registration, and Kundu et.
al.~\citep{kundu2018discovery} formulated a DBM approach, TrBM, which replaces
non-linear warps by optimal transport plans.

\section{Unbalanced Optimal Transport Morphometry}
\label{sec:methods}
The OTF approach follows the standard voxel-wise morphometry pipeline with a few
additions. The main difference is that the OTF analysis
\rev{uses}
% is a voxel-wise statistical analysis on
mass allocation images derived from the solution of unbalanced
optimal transport between the participants and a template. \rev{The basic pipeline
consists of spatial alignment, construction of a template image, computation of optimal transport features for each subject, and voxel-wise correlation of those features.}
%\begin{enumerate}
% \item Preprocessing (brain extraction, bias correction, etc.).
% \item Spatial alignment.
% \item Segmentation and construction of tissue density images.
% \item Construction of the template image.
% \item Computation of optimal transport maps for each subject.
% \item Construction of mass allocation and transport cost images.
% \item Voxel-wise correlation with mass allocation and transport cost images.
%\end{enumerate}

%The pipeline presents a few modifications to the traditional VBM/TBM approach.
%\note{The modification mentioned after this point appears to be a modification
%of the approach from the 2018 ``Exploratory Population Analysis...'' paper,
%not a modification of the traditional VBM/TBM approach.
%If I am understanding this correctly, then the first sentence of this paragraph
%needs to be rephrased or moved.}
We  introduce a template based approach instead of pairwise transport maps
between all participants to construct allocation and transport cost images. The
template-based approach (Section~\ref{sec:template}) reduces the computational
burden and restricts mass allocation to the support of the template. Finally,
we describe a modification of the unbalanced optimal transport linear program
that allows for localized mass-balancing strategies
(Section~\ref{sec:unbalanced}) and shows that OTF analysis forms a continuum
between voxel-wise VBM/TBM and a global mass balancing approach.

\subsection{Template Construction}
\label{sec:template}
For the template-based approach we considered three approaches to construct a
template from a set of spatially aligned images $X_i$. A simple approach to
construct a template is the voxel-wise mean, i.e. a Euclidean mean $E =
\frac{1}{n} \sum_i X_i$. This leads to a template with a large number of
non-zero voxels and increases the computation time of the optimal transport
described in Section~\ref{sec:unbalanced}. To reduce the number of non-zero
voxels we considered a sparse mean $E_s = E \otimes B_s$, where
$\otimes$ denotes a voxel-wise product of images, and where $B_s$ is a
binary image with
\begin{equation}
B_s(x) =
   \begin{cases}
           1, \text{if} \sum_i \mathbbm{1}_{>0}( X_i(x) ) \ge s \\
           0, \text{otherwise}.
   \end{cases}
\end{equation}
Here $\mathbbm{1}_{>0}$ is the indicator function that takes the value $1$
on positive inputs and vanishes otherwise.
In other words,
$E_s$ is the mean but only at voxels where more than $s$ images have a
positive value.

Finally, we consider the optimal transport barycenter:
\begin{equation}\label{eq:ot_barycenter}
E_o = \arg \min_Y \sum_i d(X_i, Y)^2
\end{equation}
where $d(X_i, Y)$ is the unbalanced optimal transport
distance described in Section~\ref{sec:features}.
There are several
algorithms to compute the optimal transport
barycenter~\citep{anderes2016discrete,cuturi2014fast}. We use an approximate
approach by iteratively updating the locations in $Y$ depending on the optimal
transport plans from each $X_i$.

In practice we did not see any differences in the results with different types
of means and use the sparse Euclidean mean approach due to its computational
efficiency.

\subsection{Unbalanced Optimal Transport with Local Mass-Balancing}
\label{sec:unbalanced}
For two probability measures $\mu$ and $\nu$ on probability spaces ${\Xsp}$
and ${\Ysp}$ respectively, a \emph{coupling} of $\mu$ and $\nu$ is a measure
$\coupling$ on ${\Xsp}\times{\Ysp}$ such that the marginals of $\coupling$ are
$\mu$ and $\nu$. The coupling $\coupling$ defines a {\em transport plan} that
captures how much mass $\coupling(x, y)$ is transported from $x \in \Xsp$
to $y \in \Ysp$. To define \emph{optimal} transport, we need
a cost function $\cost(x,y)$ on ${\Xsp}\times{\Ysp}$, representing the work or
cost of moving a unit of mass (tissue) from $x$ to $y$.
The cost is typically
the Euclidean distance between the tissue locations but can be defined
differently depending on the application domain, e.g. distance on the cortical
surface for surface-based analysis. An optimal coupling $\coupling^*$
minimizes this cost over all choices of couplings
$\mathcal{C}(\mu,\nu)$ between
$\mu$ and $\nu$:
\begin{equation}
  \coupling^*= \argmin_{\coupling\in\mathcal{C}(\mu,\nu)} \int_{{\Xsp}}\int_{{\Ysp}}
\cost(x,y)  d\coupling(x,y) \,.
\end{equation}

\def\nucoeff{z}

For discrete distributions $\mu = \sum_1^n w(x_i) \delta(x_i)$ and $ \nu =
\sum_1^m \nucoeff{}(y_i) \delta(y_i)$ with $\sum w(x_i) = \sum \nucoeff{}(y_i) = 1$ the optimal
transport problem can be solved by the linear program:
\begin{multline}
\min_\coupling \sum_{\substack{i=1,\dots,n\\ j=1,\dots,m}}
      \cost(x_i, y_j) \coupling(x_i, y_j) \\\
\text{s.t.}\,
\begin{cases}
  \sum_j \coupling(x_i, y_j) = \mu(\{x_i\}) = w(x_i) & \\
  \sum_i \coupling(x_i, y_j) = \nu(\{y_j\}) = \nucoeff{}(y_j) & \\
  \coupling(x_i, y_j) \ge 0
\end{cases}
\label{eq:ot}
\end{multline}

To extend the formulation to deal with positive measures $\mu$ and $\nu$ with
total mass $\lvert \mu \rvert = \sum_1^n w(x_i)$ and $\lvert \nu \rvert =
\sum_1^m \nucoeff{}(y_j)$ that can have a mass imbalance  $\Delta =  \lvert
\nu \rvert  - \lvert \mu \rvert  $, the linear program is modified
to allow for the addition and removal of mass.
%There are several possible
%strategies to include the allocation or removal of mass. One approach is
%to add $\Delta$ mass to the distribution with less mass. The second
%approach is to remove $\Delta$ mass from the distribution with more
%mass.
%\note{What does it mean to add or remove mass from
%the distrubution ``globally''? Does it mean to make the allocation
%uniform across the image? If so then are these two approaches really different?}
%Both of these approaches balance the mass globally.
One approach is to correct the mass imbalance by
adding or removing a net amount $\Delta$
uniformly across the images.
This can balance the mass \emph{globally},
but a more flexible
approach is to allow for the balancing of mass at a \emph{local} scale. This can be
important if the effect sought after is local to one region of the brain and
smaller than the overall global variation in mass.  In this case the local
effect would be masked due to the larger global variation. We propose an
approach that provides a continuum between 
local VBM/TBM and the global OTF
approach by allowing for both removal and addition of mass and including a cost
for adding or removing mass.
When the mass allocation cost is zero, the minimum cost plan ends up involving only mass allocation and no mass transport, effectively reducing the features
to those of the classical VBM/TBM approach.
By including a cost to the allocation and removal
of mass, a trade-off is made between moving mass between locations or adding
and removing mass.  Allocation or removal of mass happens only if the cost of
transporting mass to a given location is higher than the allocation cost. This
mass-balancing optimal transport problem 
can be formulated as follows:
%\endnote{The final two terms in
%equation (\ref{eq:unbalanced-obj}) are newly added; they were previously
%omitted in error. These source-sink connections are needed
%to allow for mass balancing with constraints on mass added/removed for unequal measures.}
\begin{align}
(\tau^*,\pi^*) \ =\  &\argmin_{\allocation, \coupling}
\sum_{i=1,\dots,n} \sum_{j=1,\dots,m} \cost(x_i, y_j) \coupling(x_i, y_j)\ +\nonumber\\
&\sum_{i=1,\dots,n} \left( \acost(x_a, x_i) \allocation(x_a, x_i) +
	\acost(x_r, x_i) \allocation(x_r, x_i) \right)\ +\nonumber\\
&\sum_{j=1,\dots,m} \left( \acost(y_a, y_j) \allocation(y_a, y_j) +
	\acost(y_r, y_j) \allocation(y_r, y_j) \right)\label{eq:unbalanced-obj}
,
\end{align}
where $x_a$ and $y_a$ are virtual mass allocation locations,
$x_r$ and $y_r$ are virtual mass removal locations,
$\allocation$ \rev{encodes the
amounts of mass allocation and removal
}
at cost $\acost$,
and $(\tau,\pi)$ is
\rev{constrained by}
\begin{align*}
 \sum_j \coupling(x_i, y_j) - \allocation(x_a,x_i) + \allocation(x_r,x_i) 
 	&\ =\ w(x_i)  \\
 \sum_i \coupling(x_i, y_j) + \allocation(y_a,y_j) - \allocation(y_r,y_j) 
 	&\ =\ \nucoeff{}(y_j)  \\
\rev{\sum_i(\allocation(x_a,x_i) - \allocation(x_r,x_i)) -
\sum_j( \allocation(y_a,y_j) - \allocation(y_r,y_j))} &= \rev{\Delta} \\
 \coupling(x_i,y_j), \allocation(x_a,x_i), \allocation(x_r,x_i), 
\allocation(y_a,y_i), \allocation(y_r,y_i),
&\ \ge\ 0 \\
\end{align*}
Solving the unbalanced optimal transport problem is convex and yields a global
minimum. After choosing the parameters, a transport cost $\cost$
and an allocation cost $\acost$, the unbalanced optimal transport can be solved as a linear program.
By appropriately setting the
bounds on mass allocation, mass removal, and mass exchange
this linear program can accommodate different mass-balancing schemes. The
linear program is a standard discrete optimal transport problem and can be
incorporated into fast approximation algorithms for large data sets such as the
Sinkhorn approach~\citep{cuturi2013sinkhorn} or multiscale
strategies~\citep{gerber2017multiscale}.

\subsection{Construction of Allocation and Transport Cost Images}
\label{sec:features}
For an image $X_k$ denote by $X_k(x_i)$ the associated non-negative value at
voxel location $x_i$, which defines a measure $\mu_k = \sum_{i=1}^n X_k(x_i)
\delta(x_i)$.  Similarly, given a template $T$ define the measure
$\nu = \sum_{i=1}^n T(x_i) \delta(x_i)$.
To construct the voxel-wise mass allocation
and transport cost images, we solve for optimal transport $\coupling^*_{k}$  and
allocation $\allocation^*_{k}$ plans of the unbalanced optimal transport
problem
from a fixed template $T$ to each image $X_k$.
The
variable $\allocation$ in equation (\ref{eq:unbalanced-obj}) captures the amount of
mass allocated when moving mass from $T$ to $X_k$. For image $X_k$ the
mass allocation image $M_k$ is constructed by
$$
M_k(x_i) =
  \allocation^*_k(x_a, x_i) - \allocation^*_k(y_a, x_i) -
  \allocation^*_k(x_r, x_i) + \allocation^*_k(y_r, x_i) \, .
$$
The image $M_k$ captures mass allocated in the template and mass removed in
$X_k$ as positive, and mass removed in the template and mass added in $X_k$ as
negative. The transport cost image is constructed with
$$
C_k(x_i) =
\sum_j \coupling^*_{k}( x_i, x_j ) \cost(x_i, x_j) - \sum_j
\coupling^*_{k}( x_j, x_i ) \cost(x_j, x_i) \, .
$$
The image $C_k$ captures locations that have  large transportation, i.e. large
cost, out of a template location as positive and locations that see
large transportation into $X_k$ as negative.

The images $M_k$ and $C_k$ are smoothed with a small Gaussian to increase
correlations between neighboring pixels. The smoothed images $M_k$ and $C_k$
replace the smoothed intensity or Jacobian determinant images in the
statistical analysis of a VBM or a TBM pipeline.
\rev{The final pipeline is summarized along with additional details of our approach in Section~\ref{sec:anameth}.}

\begin{figure}[tbh]
\centering
\renewcommand\tabcolsep{0.15mm}
\renewcommand\arraystretch{0.5}
\begin{tabular}{l|ccccc}
  \multirow{3}{*}{ \raisebox{0mm}[16mm][0mm]{\includegraphics[width=0.078\linewidth]{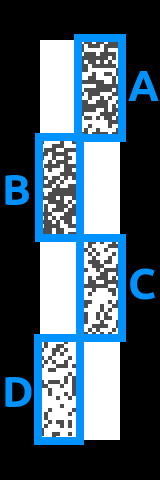} } }&
  \raisebox{2mm}[0mm][0mm]{ (a) } &
\includegraphics[width=0.065\linewidth, angle=90]{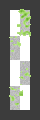} &
\includegraphics[width=0.065\linewidth, angle=90]{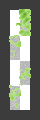} &
\includegraphics[width=0.065\linewidth, angle=90]{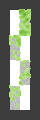} &
\includegraphics[width=0.065\linewidth, angle=90]{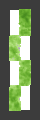} \\
  & \raisebox{2mm}[0mm][0mm]{ (b)}&
\includegraphics[width=0.065\linewidth, angle=90]{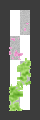} &
\includegraphics[width=0.065\linewidth, angle=90]{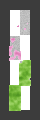} &
\includegraphics[width=0.065\linewidth, angle=90]{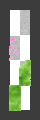} &
\includegraphics[width=0.065\linewidth, angle=90]{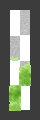} \\
  & \raisebox{2mm}[0mm][0mm]{ (c)} &
\includegraphics[width=0.065\linewidth, angle=90]{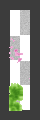} &
\includegraphics[width=0.065\linewidth, angle=90]{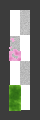} &
\includegraphics[width=0.065\linewidth, angle=90]{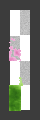} &
\includegraphics[width=0.065\linewidth, angle=90]{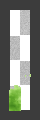} \\
  & &\multicolumn{4}{c}{local ($\acost=0$)   $\xlongleftrightarrow{\hspace{20mm}}$  global ($\acost=\infty$)} \\
  (I)   & \multicolumn{5}{l}{\,(II) }\\[2mm]
  &\multicolumn{5}{c}{\,-0.65 \includegraphics[width=60mm, height=2.5mm, angle=0]{colorbar} 0.65}
\end{tabular}
\caption{\label{fig:strip-localize}
\rev{A toy example simulating tissue deterioration to illustrate the
behaviour of the continuum from local to global mass-balancing. (I)
Input image example; each of the regions (A,B,C,D) has various random amounts
of white pixels removed, representing tissue deterioration.
(II) Suppose that there is a disease (a) corresponding to tissue loss in regions A+B+C+D,
a disease (b) corresponding to tissue loss in regions C+D, and a disease (c)
corresponding to tissue loss in region D.
Here we display the correlation of each disease with the mass allocation OTF at each voxel,
and we do this for various choices of $\acost$.}
The results illustrate that global mass-balancing leads to stronger correlations
compared to local mass-balancing for tissue loss over a large region.
However, if only a local region is affected as in (c) and there is
confounding (i.e.  uncorrelated) tissue loss in other regions, the
correlation strength decreases compared to the more local approaches.
}
\end{figure}

\begin{figure}[hbt]
\centering
\renewcommand\tabcolsep{0.5mm}
\renewcommand{\arraystretch}{0.75}
\begin{tabular}{cc @{\hspace{1mm}} c @{\hspace{1mm}}cc}
  \multicolumn{2}{c}{Case 1} && \multicolumn{2}{c}{Case 2} \\
  allocation & transport & & allocation & transport \\\cline{1-2}\cline{4-5}
  \\[-2mm]
\includegraphics[width=0.225\linewidth]{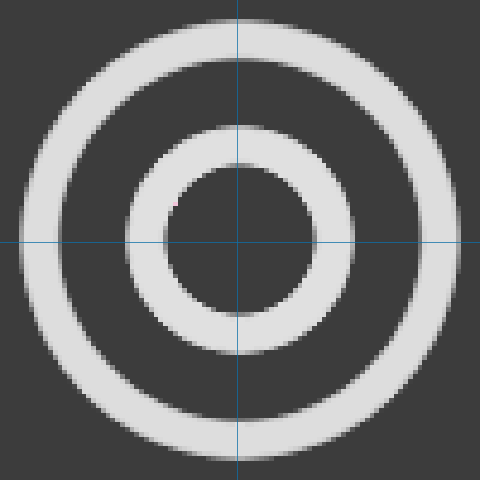} &
  \includegraphics[width=0.225\linewidth]{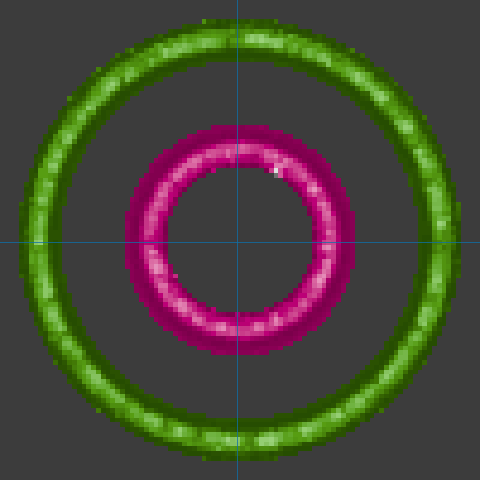} &&
  \includegraphics[width=0.225\linewidth]{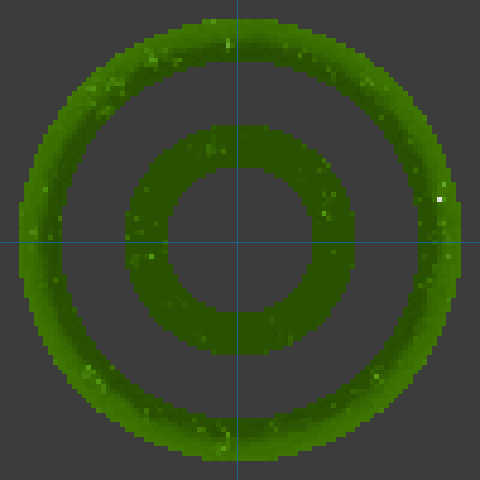} &
\includegraphics[width=0.225\linewidth]{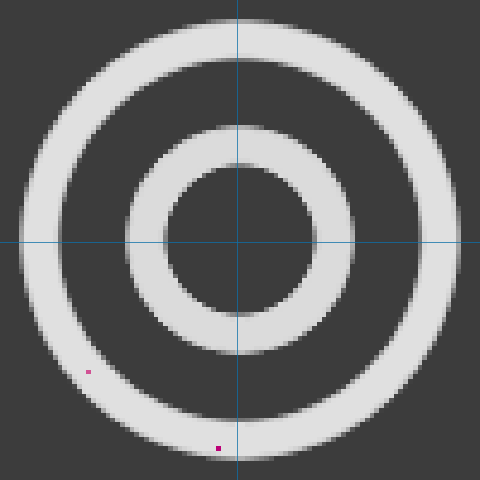} \\
\\
\multicolumn{5}{c}{-0.65 \includegraphics[width=60mm, height=2.5mm, angle=0]{colorbar} 0.65}
\end{tabular}
\caption{\label{fig:allocation-transport}
\rev{Correlation to allocation and transport features in two cases on a toy example.
In Case 1,
each sample image has the same overall mass, but the mass is differently distributed
among the two annuli.
Color shows correlation between the total mass of the outer annulus and the OTF (mass allocation and transport cost) at each pixel.
We see that OTF correctly identified that the effect is due to a trade-off in mass
between the inner and outer annuli,
and not due to random variation of total mass.}
There is a statistically significant correlation to
transport, but not to allocation.
In Case 2, each sample image has a
random overall mass which is randomly distributed between the two annuli.
Color shows correlation between the total mass of both annuli and the OTF at each pixel.
We see that OTF correctly \rev{identified} the overall difference in mass.
}
\end{figure}

\rev{Figure~\ref{fig:strip-localize} uses a toy example to show the effects of
different mass-balancing strategies (different choices of $\acost{}$)
and different types of tissue loss
on the sensitivity of OTF.
The example consists of 20 2D images with four rectangular patches
that have different amounts of missing tissue.
}
\rev{Figure~\ref{fig:allocation-transport}
uses another toy example to illustrate
that the transport and
allocation features can differentiate effects due to change in the location of
mass from effects due to overall mass.
The example consists of 40 2D
images with two concentric annuli
that have different amounts of mass (gray scale
intensity).}

In addition to defining transport and allocation images,
optimal transport may also be used to define a notion of distance between two images.
The \emph{unbalanced optimal transport distance} between two images is defined to be the minimum value achieved
by the objective on the right hand side of (\ref{eq:unbalanced-obj}).
This provides an interesting alternative way to construct a template image, described in (\ref{eq:ot_barycenter}).

\section{Statistical Analysis}
\label{sec:analysis}
This section provides an analysis of the asymptotic correlation strength
improvement of OTF over VBM for spatially dispersed effects.
We analytically derive
Pearson's correlation coefficients for a pathology scenario with dispersed
tissue loss with and without the OTF, showing that OTF
significantly improve correlation strength.  In Section~\ref{sec:smoothing} we
illustrate the results of the analysis on a toy example that demonstrates the
effect of sample size and smoothing on correlation strength.

\subsection{Analysis of Correlation Strength}
\label{sec:analysis1}
Consider a study with participants that  have a pathology with
probability $p$ and are healthy with probability $1-p$. We encode this in the
random variable $H$ with Bernoulli distribution: the probability of observing a
healthy participant is $P(H = 1) = 1-p$ and
the probability of observing
a pathological participant is
$P(H=0) = p$.  The effect of the pathology is a tissue loss in some region of
the brain. We model the region here as $n$ discrete locations, e.g. the voxels
of a brain MRI. We model tissue loss as follows:
the probability of observing tissue at
a location $X$ is $P(T_X = 1 | H = 1) = t_h$ for the healthy population and $P(T_X = 1 | H = 0 ) = t_p$ for the pathological population.
We assume that these probabilities do not vary with $X$; i.e., they are the same for each of the $n$ voxel locations.

The correlation between pathology indicator $H$ and tissue
expression $T_X$ is
\begin{equation}
  \label{eq:cor}
        \cor(T_X, H) = \frac{E[ T_X H ] - E[T_X]E[H] }{ \sigma^2(T_X) \sigma^2(H) }
\end{equation}
with $E[]$ the expectation and 
$\sigma^2(T_X) = E[T_X^2] - E[T_X]^2$ the variance. The
correlation $\cor(T_X, H)$ is asymptotically the correlation estimated by VBM at
voxel $X$.  The correlations for the VBM approach
$\cor(T_X, H)$, which may be computed from
the distribution of the population $P(H)$ and the conditional tissue loss
probabilities $P(T_X = 1 | H = 1 ) = t_h$ and $P(T_X = 1 | H = 0 ) = t_p$.

OTF consist of two features at each voxel: the transport cost image and the mass allocation image. We will focus on the latter and assume that there is no transport, as is the case in the limit of \emph{local} mass balancing, where $\acost{}\ll\cost{}$. Determining the asymptotic correlation strength for OTF requires computing
the probability of allocating or removing tissue amount $A_X$ at a location $X$ for
both the healthy and the pathological population:
$P(A_X | H=1)$ and $P(A_X | H=0)$.
We compute the allocation and removal probabilities with respect to a template
that is constructed by averaging the healthy population:
$T(X) = E[T_X|H=1] = t_h$.
For the analysis assume that the OTF exclusively allocate or
remove tissue and never combine these operations at an individual location.  Then the probability of 
allocating $A_X=a$ units of tissue
is zero 
except when $a=-t_h$ or $a=1-t_h$;
i.e., either $t_h$ tissue is removed or $1 - t_h$ tissue is added.
The distribution $A_X$ depends on the total tissue $k$ of the participant.
For $k >  n t_h$ the probability of allocating tissue at 
the location $X$ is
$\frac{ k - n t_h}{ n (1 - t_h) }$ and for $k < n t_h$ the probability of
removing tissue is $\frac{ t_h n - k }{n t_h}$.
The probability of a
participant having tissue at $k$ locations,  $P(M=k)$, is binomial distributed
with $n$ trials and probability of success $t_h$ and $t_p$ for the healthy and
pathological participants, respectively, i.e.  $M|(H=1) \sim B(n, t_h)$ and
$M|(H=0) \sim B(n, t_p)$. 
With this model we can compute the probability of
removing or allocating tissue as:

\begin{small}
\begin{flalign*}
  &P(A_X = -t_h | H=1) = \sum_{k=1}^{\floor*{t_hn}}   P(M = k | H = 1) \frac{ t_h n - k }{n t_h}    \\
  &P(A_X = 1- t_h | H=1) = \sum_{k=\ceiling*{t_hn}}^n P(M = k | H = 1) \frac{ k - t_h n }{n (1-t_h)} \\
  &P(A_X = -t_h | H=0) = \sum_{k=1}^{\floor*{t_hn}}   P(M = k | H = 0) \frac{ t_h n - k }{n t_h}     \\
  &P(A_X = 1-t_h | H=0) = \sum_{k=\ceiling*{t_hn}}^n P(M = k | H = 0) \frac{ k - t_h n }{n (1-t_h)}
\end{flalign*}
\end{small}
The expectations $E[A_XH]$, $E[A_X]$ and $E[A_X^2]$ needed to compute the correlation
$\cor(A_X, H)$ are readily obtained.

\begin{figure}[htb]
%\scriptsize
\centering
\renewcommand\tabcolsep{0.15mm}
\renewcommand\arraystretch{1}
\begin{tabular}{cc}
\includegraphics[width=0.49\linewidth]{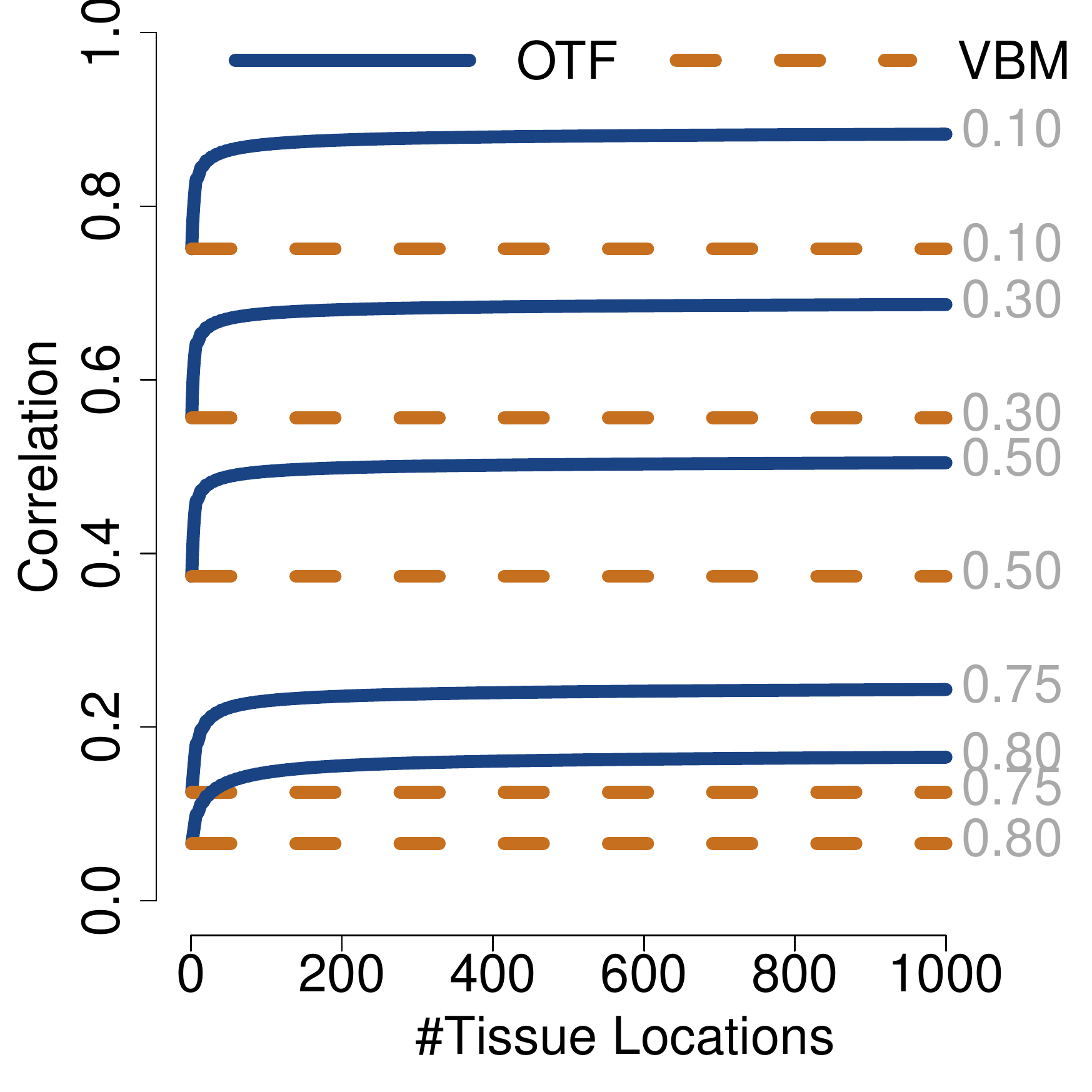} &
\includegraphics[width=0.49\linewidth]{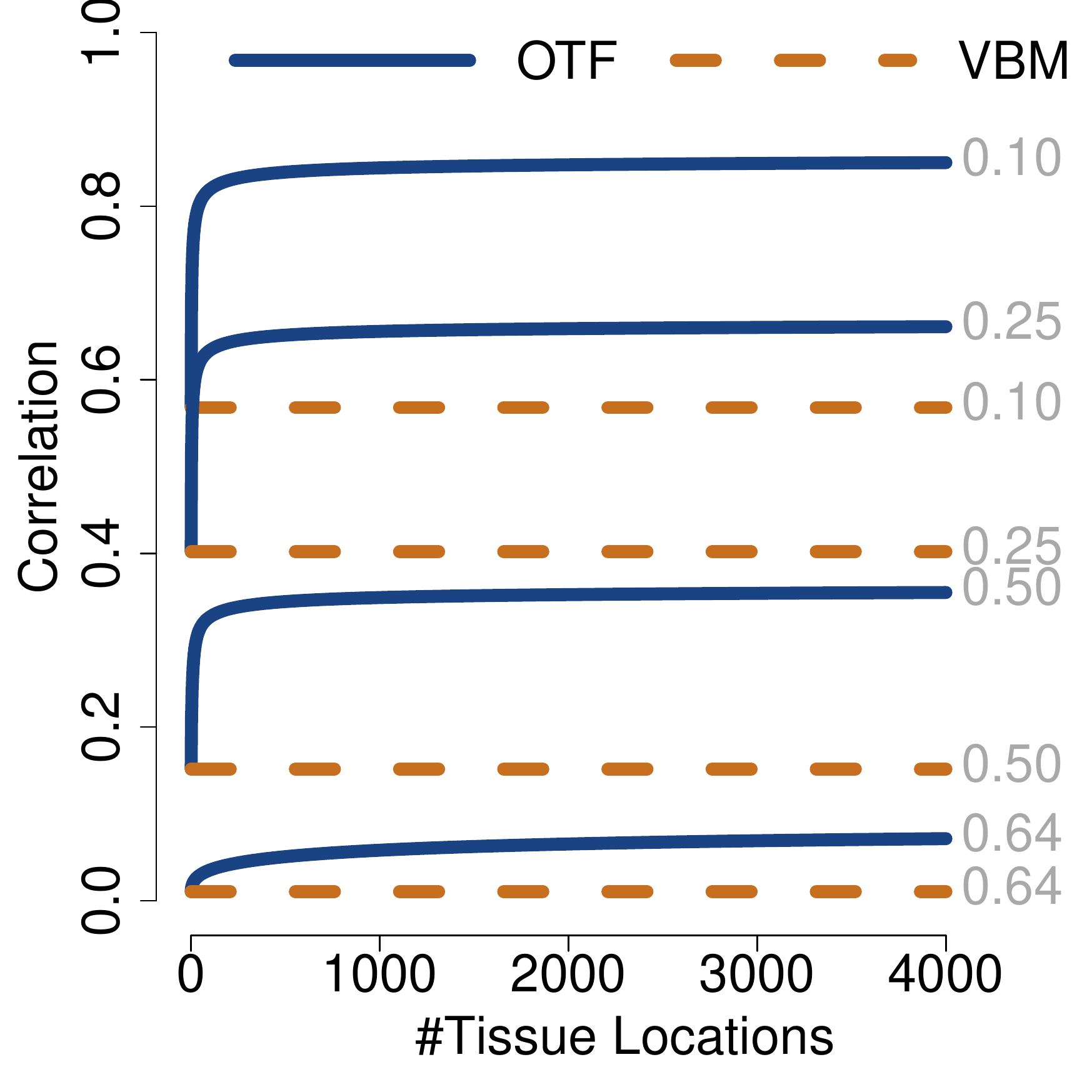} \\
        % (a) & (b) \\
        (a) {\scriptsize $t_h = 0.85$ } &(b) {\scriptsize  $t_h = 0.65$} \\
        {\scriptsize  $t_p = 0.1, 0.3, 0.5, 0.75, 0.8$} & {\scriptsize  $t_p = 0.1, 0.25, 0.5, 0.64$ }
\end{tabular}
\caption{\label{fig:stats2}
Plots of the correlation strength for the analysis described in the text.
Correlation as a function of the number of tissue locations with  (a) $t_h =
0.85$ and (b) $t_h=0.65$ and varying $t_p$ for $p=0.5$. The OTF extraction step
improves correlation for dispersed tissue loss: An increasing number of
locations with tissue loss increases the correlation strength when using OTF.}
\end{figure}

Figure~\ref{fig:stats2} illustrates the effect of OTF on asymptotic
correlation strength by plotting 
$\cor(T_X, H)$ and $\cor(A_X, H)$
as a function of
the number of tissue locations $n$, i.e. how dispersed the pathology is, for
varying parameters.
Including OTF in the voxel-wise morphometric analysis increases correlation
strength with increasing number of tissue locations. With increasing number of
locations, the expected total tissue difference is increased resulting in more
locations that require tissue removal. On the other hand, if there are local
variations in the populations independent of total tissue loss or gain,
then OTF can decrease in statistical power.

\subsection{Effects of Smoothing and Sample Size}
\label{sec:smoothing}
\begin{figure}[hbt]
\centering
\renewcommand\tabcolsep{0.15mm}
\renewcommand{\arraystretch}{0.25}
\begin{tabular}{llcccc}
  & & $n=20$ & $n=40$ & $n=80$ & $n=160$ \vspace{1mm} \\

\multirow{2}{*}{ \raisebox{1mm}[0mm][0mm]{\rotatebox[origin=c]{90}{$\sigma=0$}} }&
        \raisebox{2mm}{\rotatebox[origin=c]{90}{V}} &
\includegraphics[width=0.075\linewidth, angle=90]{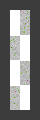} &
\includegraphics[width=0.075\linewidth, angle=90]{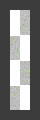} &
\includegraphics[width=0.075\linewidth, angle=90]{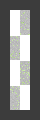} &
\includegraphics[width=0.075\linewidth, angle=90]{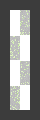} \\
\vspace{1mm}
  &\raisebox{2mm}{\rotatebox[origin=c]{90}{O}} &
\includegraphics[width=0.075\linewidth, angle=90]{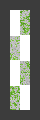} &
\includegraphics[width=0.075\linewidth, angle=90]{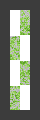} &
\includegraphics[width=0.075\linewidth, angle=90]{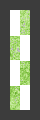} &
\includegraphics[width=0.075\linewidth, angle=90]{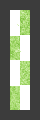} \\

\multirow{2}{*}{ \raisebox{1mm}[0mm][0mm]{\rotatebox[origin=c]{90}{$\sigma=1$}} }&
\raisebox{2mm}{\rotatebox[origin=c]{90}{V}} &
\includegraphics[width=0.075\linewidth, angle=90]{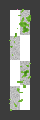} &
\includegraphics[width=0.075\linewidth, angle=90]{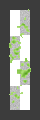} &
\includegraphics[width=0.075\linewidth, angle=90]{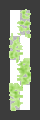} &
\includegraphics[width=0.075\linewidth, angle=90]{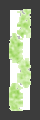} \\
\vspace{1mm}
  &\raisebox{2mm}{\rotatebox[origin=c]{90}{O}} &
\includegraphics[width=0.075\linewidth, angle=90]{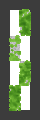} &
\includegraphics[width=0.075\linewidth, angle=90]{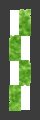} &
\includegraphics[width=0.075\linewidth, angle=90]{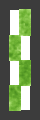} &
\includegraphics[width=0.075\linewidth, angle=90]{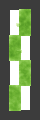} \\

\end{tabular}
\caption{\label{fig:strip-smoothing}
\rev{Effects of smoothing and sample size on correlation of features to total mass.
The sample images are generated as in Figure~\ref{fig:strip-localize},
and we look at two types of features: VBM (V), which considers voxel intensities, and OTF (O) which produces a mass allocation image out of a solution to the unbalanced optimal transport problem.}
The correlation strength is indicated in green at locations with statistical
significance $p<0.05$.  VBM requires a large amount of smoothing ($\sigma$)
and more samples ($n$) to find statistically significant correlations and
results in weaker correlations than OTF with less smoothing and fewer samples.
However,
OTF with large smoothing can result in a loss of spatial resolution, which
leads to the inclusion of
large amounts of healthy tissue.}
\end{figure}

\rev{The VBM method requires smoothing in order to convert a tissue mask into a measure of tissue concentration.
The smoothing kernel for VBM needs to be large in order to increase the dependency of neighboring voxels and capture effects that vary in their spatial location.
While TBM does not need smoothing, it is helpful in terms of statistical sensitivity, and a large smoothing kernel becomes necessary as well if one wants to capture spatially varying effects.}
%The VBM/TBM methods recommend a relatively large amount of smoothing to
%increase dependency of neighboring voxels and thus reduce sensitivity to
%spatial location of the effects.
The OTF method increases statistical power for diffuse
tissue loss and can mitigate effects caused by shifts. However, the exact
location for mass allocation is still variable for each image. Thus, the OTF analysis should
still employ smoothing to increase spatial dependencies, but
it should do so after the
computation of the optimal transport solutions. \rev{Smoothing} is applied
to the mass allocation and transport cost images and not to the input images.
At this stage the smoothing has the same effect as smoothing for VBM and
increases correlation strength at the cost of spatial resolution.
\rev{Figure~\ref{fig:strip-smoothing} empirically illustrates the interaction of
sample size and smoothing for OTF and VBM on the
toy example from Figure~\ref{fig:strip-localize}.}

VBM requires a significant amount of smoothing for small to moderate sample
size to discover diffuse tissue loss effects. Using OTF detects disperse
tissue loss for small samples with little smoothing. We suggest using a
truncated Gaussian, or any other kernel with limited range, in order to limit
the spatial extent to which signals can be correlated due to smoothing.  For
large amounts of smoothing, even with a truncated kernel, the smoothing can
yield false positive correlations, as shown in Figure~\ref{fig:strip-smoothing}
where the effects leak into the healthy white regions.

\section{Results}
\label{sec:results}
We demonstrate the effect of OTF on the OASIS-1 data. The OASIS-1 data set was
collected for studying the effects of cognitive impairment. The OASIS-1
database consists of T1-weighted MRI and tissue segmentation masks of 416
participants aged 18 to 96~\citep{marcus2010open}. The images in the OASIS-1
data set are brain-extracted, gain-field corrected, and registered to the
Talaraich atlas space~\citep{talaraich:book88} with an affine
transform.

\begin{figure}[tb]
\small
\setlength{\tabcolsep}{0pt}
\renewcommand{\arraystretch}{1.2}
\centering
  \begin{tabular}{ccc}
          \hspace{6mm} Healthy &  \hspace{6mm} Very Mild & \hspace{6mm}  Mild\\
  \includegraphics[width=0.32\linewidth]{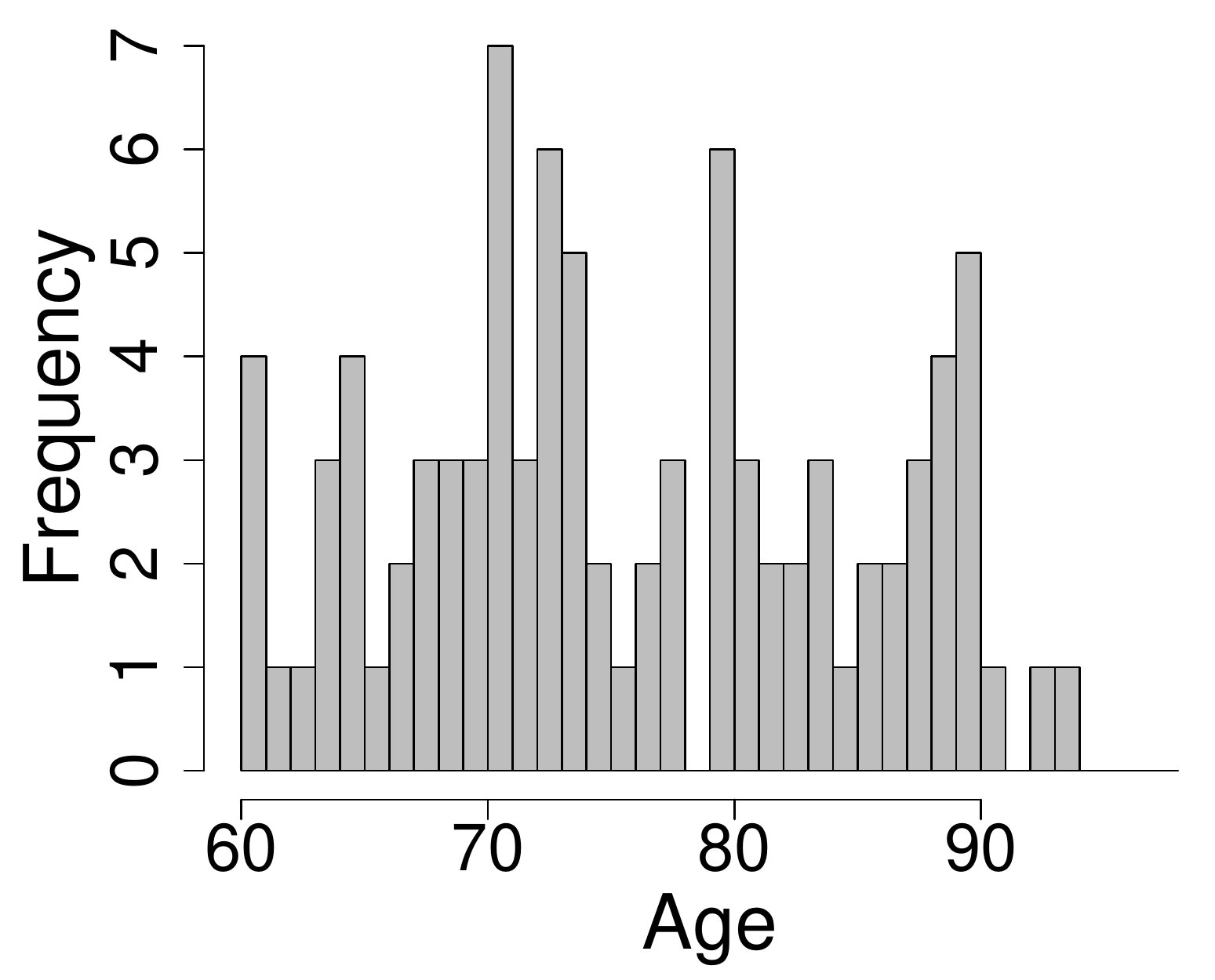} &
  \includegraphics[width=0.32\linewidth]{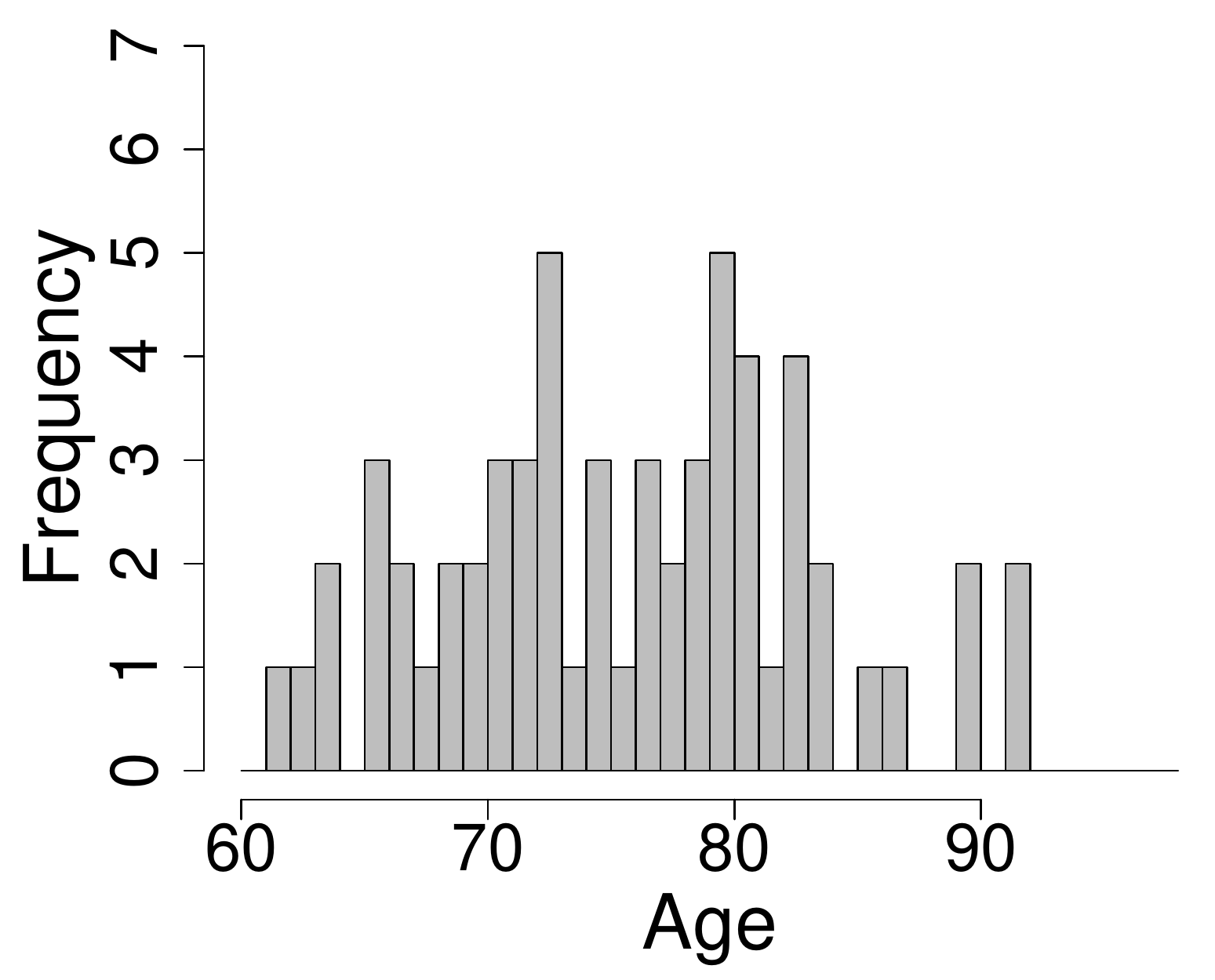} &
  \includegraphics[width=0.32\linewidth]{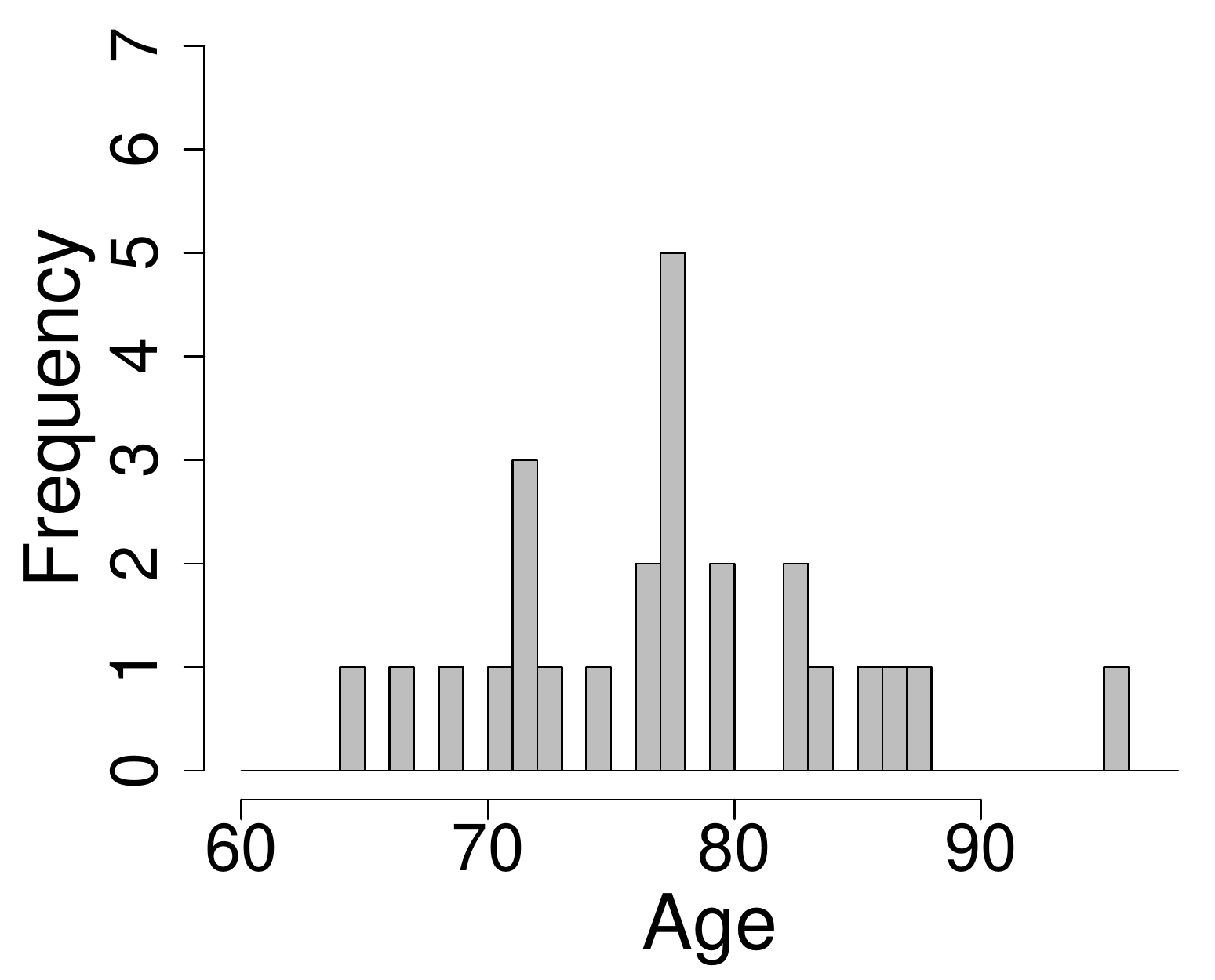} \\
\end{tabular}
\caption{\label{fig:oasis-age-hist}
Histogram of ages per CDR group of the OASIS-1 data.
}
\end{figure}

The OASIS-1 data set has a clinical dementia rating (CDR) associated to each
participant.  We restrict the analysis to 177 participants aged sixty and
over to reduce the effects of age. The clinical dementia rating of the 177
participants falls into 3 categories: healthy (CDR  = 0, 90 participants), very
mild (CDR = 0.5, 60 participants), mild  (CDR = 1, 25 participants), and
moderate dementia (CDR = 2, 2 participants).  Figure~\ref{fig:oasis-age-hist}
shows histograms of age for each CDR rating of the population used in this
analysis.

\subsection{Analysis Methodology}
\label{sec:anameth}

\def\vbmaff{VBM-Aff}

To explore the effect of the optimal transport feature extraction in both the
VBM and TBM setting we registered the T1-weighted images to the SRI24
atlas~\citep{rohlfing2010sri24} using an affine registration followed by {SyN diffeomorphic registration}~\citep{avants2008symmetric}, a
state-of-the-art non-parametric registration method. For the VBM analysis we align the tissue
segmentations to SRI24 with the affine registration only; \rev{we will refer to this as \vbmaff{}}. For the TBM analysis
we align the tissue segmentations to SRI24 with an affine plus nonlinear warp
and multiply the segmentation masks by the Jacobian determinant of the
nonlinear warp to correct for volume changes~\citep{ashburner2007fast}.
\rev{The final pipeline for our TBM analysis can be summarized as follows:
\begin{enumerate}
\item Preprocess (brain extraction and bias correction).
\item Register each image to the SRI24 atlas via SyN.
\item Segment each aligned image to obtain a gray matter mask.
\item Take the Jacobian determinant of the deformation that aligned each image and restrict it to the gray matter mask. The resulting tissue density maps $X_k$ are the main input into the optimal transport analysis pipeline.
\item Construct a template $T$ out of the $X_k$ by taking either the sparse Euclidean mean or the optimal transport barycenter.
\item Solve the unbalanced optimal transport problem from $T$ to $X_k$ for each $k$. We do this by setting up the linear program~(\ref{eq:unbalanced-obj}) as a minimum cost flow problem~\citep{Ahuja:1993:NFT:137406} and taking a multiscale approach~\citep{gerber2017multiscale}.
\item Compute the OTF, i.e. the mass allocation images $M_k$ and the transport cost images $C_k$.
\item Compute the voxel-wise correlation of $M_k$ and $C_k$ with a target variable of interest, creating correlation images like the ones seen in Figures \ref{fig:illustration}, \ref{fig:strip-localize}, \ref{fig:allocation-transport}, \ref{fig:strip-smoothing}, \ref{fig:vbm-tbm}, \ref{fig:cor-oasis-mass-balancing}, \ref{fig:dementia}, and \ref{fig:aging}.
\end{enumerate}
}

\renewcommand{\AgeH}{\rev{Age\textsubscript{N}}}

We apply VBM and TBM analysis to gray matter tissue masks with respect to
several different variables and subpopulations:
\begin{itemize}
   \item \AgeH: Age for participants with CDR $= 0$, to explore the effects of
         \rev{normal} aging.
   \item \AgeD: Age for participants with CDR $> 0$, to explore the effects of
         aging in the cognitively impaired population. CDR and age for this
         group have a weak correlation of 0.18 with a p-value of 0.13,
         suggesting that any effects discovered in this group are not related
         to CDR rating.
   \item CDR: CDR rating on all participants, to explore the overall effect of
         cognitive impairment.
   \item \CDRvm: Binary variable of CDR 0 versus CDR 0.5 participants, to
         explore the effects of cognitive impairment at a very early stage.
   \item \CDRm: Binary variable of CDR 0 versus CDR 1 participants, to explore
         the effects of the progression of cognitive impairment by comparing to
         the CDR 0.5 group.
\end{itemize}
In all figures we show
correlation strength
at all locations with Bonferroni
corrected $p < 0.05$.

\renewcommand{\VBMA}{\vbmaff{}\textsubscript{A}}
\renewcommand{\VBMT}{\vbmaff{}\textsubscript{T}}

We constructed mass allocation and transport cost images for both the 
TBM and \rev{\vbmaff{}} approach
which we will refer to as \TBMA, \TBMT~and \rev{\VBMA, \VBMT,} respectively.

The mass allocation and transport cost images are obtained as described in
Section~\ref{sec:features} by computing transport maps to a sparse Euclidean
mean template. The template is constructed
as described in Section~\ref{sec:template} using $s = 0.9n$,
where $n$ is the number of images in the data set.
For the brain images, the
optimal transport mean and the sparse Euclidean mean resulted in small
differences and did not affect the qualitative results.
For the transport cost $\cost$ we use the squared Euclidean distance between
the voxel locations to introduce a preference of many small mass
transfers
over fewer larger mass transfers.
To reduce computation costs the images are
downsampled by a factor of 
2, leading to a resolution of $88 \times 104 \times 88$ voxels with a voxel size of
$ 2 \times 2 \times 2$ mm.
The downsampling resulted
in computation times of approximately one to two hours 
per optimal
transport problem when using a multiscale optimal transport
solver~\citep{gerber2017multiscale}.\rev{}

Software to replicate the results can be found on
github\footnote{\url{https://github.com/KitwareMedical/UTM}}. The github
repository also contains links to interactive web visualizations of the results
presented here.

\subsection{\rev{Non-linear TBM and Affine Only VBM Analysis}}
\begin{figure}[tbh]
\footnotesize
\setlength{\tabcolsep}{0.5pt}
\renewcommand{\arraystretch}{1.2}
\centering
\begin{tabular}{ccc @{\hspace{1mm}} c @{\hspace{1mm}} ccc}
  \multicolumn{3}{c}{TBM} & \hspace{1mm}& \multicolumn{3}{c}{\TBMA } \\
  \includegraphics[height=0.15\linewidth]{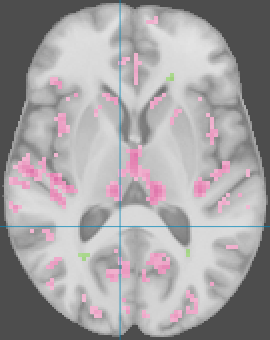} &
  \includegraphics[height=0.15\linewidth]{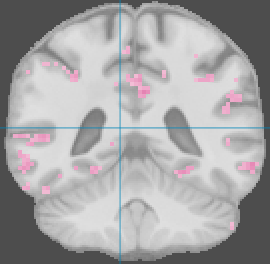} &
  \includegraphics[height=0.15\linewidth]{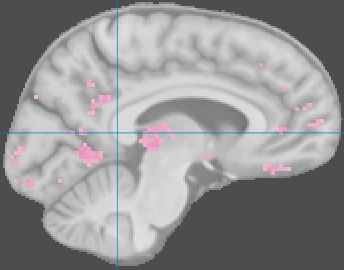} & &
  \includegraphics[height=0.15\linewidth]{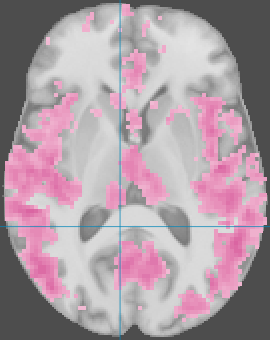} &
  \includegraphics[height=0.15\linewidth]{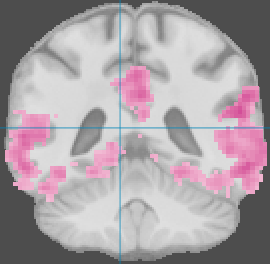} &
  \includegraphics[height=0.15\linewidth]{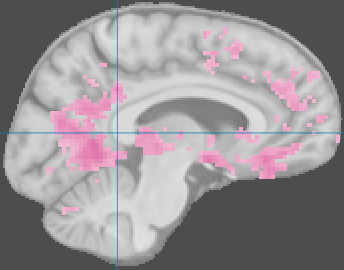} \\[2mm]
  \multicolumn{3}{c}{\rev{\vbmaff{}}} & \hspace{1mm}& \multicolumn{3}{c}{\rev{\VBMA} } \\[1mm]
  \includegraphics[height=0.15\linewidth]{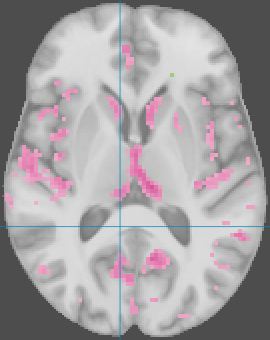} &
  \includegraphics[height=0.15\linewidth]{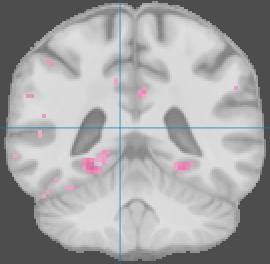} &
  \includegraphics[height=0.15\linewidth]{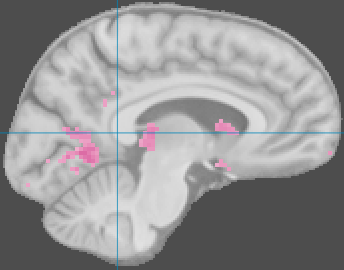} & &
  \includegraphics[height=0.15\linewidth]{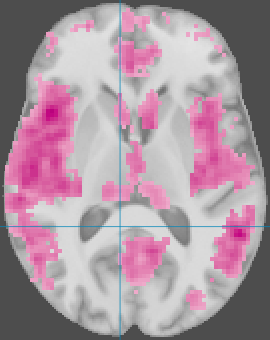} &
  \includegraphics[height=0.15\linewidth]{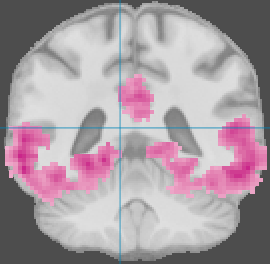} &
  \includegraphics[height=0.15\linewidth]{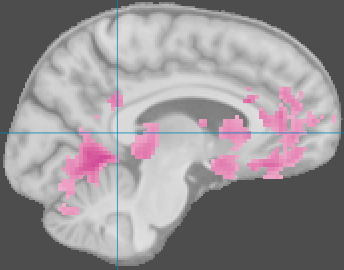} \\

  \multicolumn{7}{c}{-0.65 \includegraphics[width=60mm, height=2.5mm, angle=0]{colorbar} 0.65}
\end{tabular}
\caption{\label{fig:vbm-tbm}
TBM and VBM analysis of gray matter tissue \rev{without and with} the optimal transport
feature extraction step. The color shows \rev{inverse} correlation strength \rev{(pink)} to clinical
dementia rating for voxels with Bonferroni corrected $p < 0.05$.
}
\end{figure}

Figure~\ref{fig:vbm-tbm} illustrates the effects of using TBM and \rev{} with and
without OTF.
%Using affine versus  non-parameteric
%registration has a negligible
%effect, as previously observed in the literature on VBM and
%TBM~\cite[Chaper~6]{frackowiak2004human}. The Jacobian determinant of the
%non-parametric registration will deviate from 1 (i.e. indicate contractions or
%expansions) near tissue boundaries, which is exactly where the affine
%registration will be mismatched as well.
\rev{
The use of affine registration in the VBM case
is a simple way to ensure that local structures are not aligned
so perfectly as to wash out the signal that can be observed by VBM.
If we try to align each gyrus and sulcus,
as we do with non-parametric registration,
then it becomes more appropriate to use TBM~\cite[Chaper~7]{frackowiak2004human}.
We should expect similar results from TBM and \vbmaff{}
and similar results from \TBMA{} and \VBMA{}. This is because
the Jacobian determinant of the non-parametric registration
will deviate from $1$ (i.e. indicate contractions or
expansions) near tissue boundaries, and this is exactly where the affine
registration is expected to be mismatched.
}
Using OTF has the same effect in both
settings with stronger correlations over larger regions.
The results show the increased statistical power provided by OTF
and the capability to compensate for
shifts in exact tissue location due to registration mismatches.

\subsection{Local to Global Analysis}
\begin{figure}[hbt]
\small
\setlength{\tabcolsep}{0pt}
\renewcommand{\arraystretch}{1.2}
\centering
\begin{tabular}{ccccc}
  \multicolumn{5}{c}{TBM  (local) $\xlongleftrightarrow{\hspace{35mm}}$ (global) \TBMA} \\[1mm]

  \includegraphics[width=0.185\linewidth]{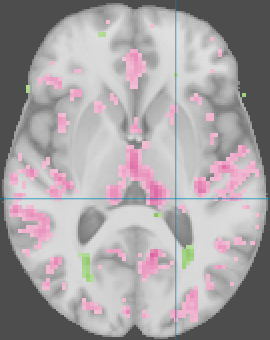} &
  \includegraphics[width=0.185\linewidth]{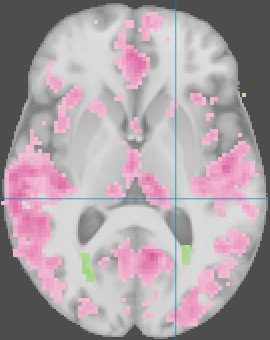} &
  \includegraphics[width=0.185\linewidth]{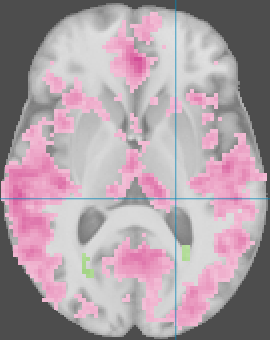} &
  \includegraphics[width=0.185\linewidth]{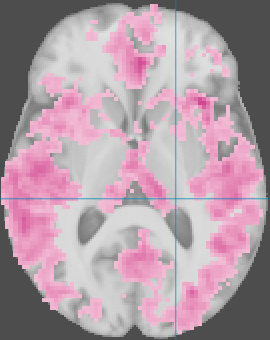} &
  \includegraphics[width=0.185\linewidth]{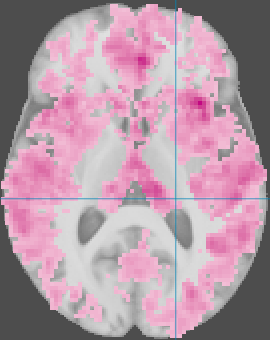} \\

  \includegraphics[width=0.185\linewidth]{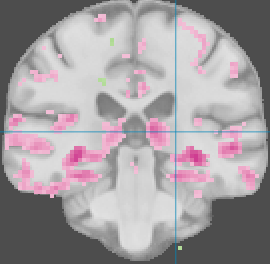} &
  \includegraphics[width=0.185\linewidth]{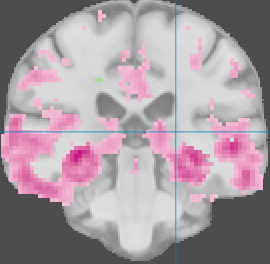} &
  \includegraphics[width=0.185\linewidth]{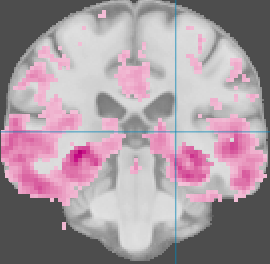} &
  \includegraphics[width=0.185\linewidth]{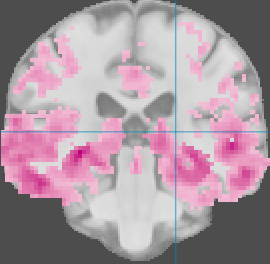} &
  \includegraphics[width=0.185\linewidth]{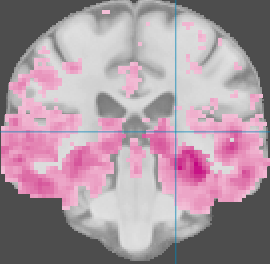} \\

  \includegraphics[width=0.185\linewidth]{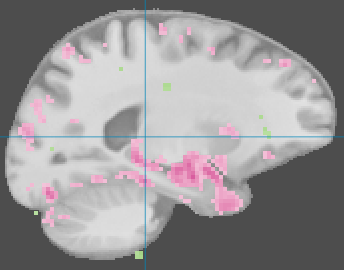} &
  \includegraphics[width=0.185\linewidth]{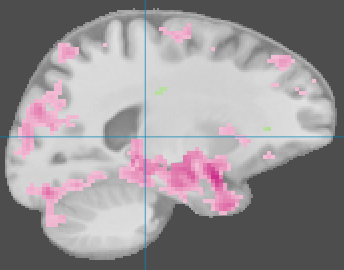} &
  \includegraphics[width=0.185\linewidth]{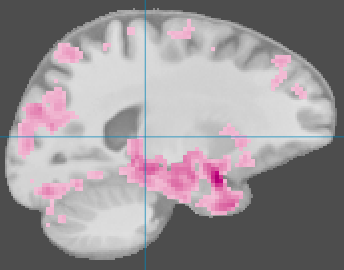} &
  \includegraphics[width=0.185\linewidth]{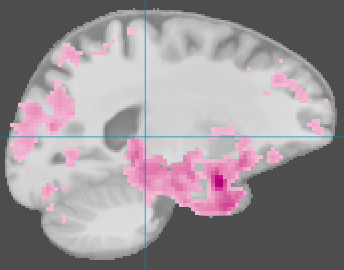} &
  \includegraphics[width=0.185\linewidth]{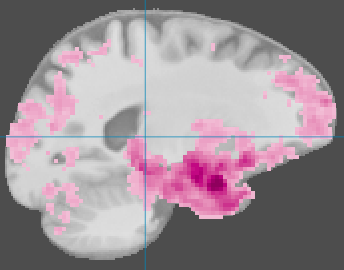} \\[1mm]

  \multicolumn{5}{c}{-0.55 \includegraphics[width=60mm, height=2.5mm, angle=0]{colorbar} 0.55}
\end{tabular}
\caption{\label{fig:cor-oasis-mass-balancing}
\rev{Inverse} correlations of gray matter to CDR, illustrating the continuum
from TBM analysis to \TBMA{} with global mass balancing.
From left to right the cost
of allocating mass is increased, causing a more global mass
balancing to be enforced.
The global \TBMA{} analysis can mask local effects if
there is no overall difference in tissue loss between the populations.
Performing the analysis at different spatial scales improves the odds of
discovering local and global effects.}
\end{figure}

Figure~\ref{fig:cor-oasis-mass-balancing} illustrates the continuum from local
(TBM) to global (\TBMA) mass-balancing strategies, as described in
Section~\ref{sec:unbalanced}, on an analysis of correlation of gray matter
amount to CDR.
The global versus local analysis has two complementary effects. For
pathologies that result in a widespread tissue loss with large individual
variation in the exact location of tissue loss, the global analysis will
increase correlation strength and potentially discover effects not apparent
with a local analysis. On the other hand, if the pathology has locally
concentrated effects, large individual variations in total volume can mask this
effect in the global analysis and a more local analysis would be needed.
Figure~\ref{fig:cor-oasis-mass-balancing} shows some of these effects.
%\note{Does it really show both of the effects?
%I'm only seeing the former effect. The only areas in the figure
%where the local analysis shows something that the global analysis misses
%are the green splotches, and these
%are described as some kind of error in the next paragraph. That
%doesn't qualify as showing the effectiveness of a more local
%analysis on locally concentrated pathologies.}
We suggest
running OTF analysis at multiple scales, as shown in
Figure~\ref{fig:cor-oasis-mass-balancing}, and comparing the results to get a
more complete picture of potential effects.

The global analysis discovers a more widespread loss of tissue (negative
correlation), particularly in the temporal lobes, not visible with a local
analysis.  The local analysis shows a tissue gain near the ventricles. A tissue
gain in the cortex is unlikely and this effect could be due to CSF appearing
brighter in participants with CDR $>$ 0 and being misclassified as gray matter.
This effect disappears with the global analysis; the overall loss of gray
matter is a much stronger effect than the localized tissue gain.

\subsection{Increase in Statistical Power}
\begin{figure}[tbh]
\footnotesize
\setlength{\tabcolsep}{0.5pt}
\renewcommand{\arraystretch}{1.2}
\centering
\begin{tabular}{l ccc @{\hspace{1mm}} c @{\hspace{1mm}} ccc}
  &\multicolumn{3}{c}{TBM} & \hspace{1mm}& \multicolumn{3}{c}{\TBMA} \\[1mm]
  \rotatebox[origin=l]{90}{\hspace{0mm} \CDRvm} \hspace{0.5mm} &
  \includegraphics[height=0.145\linewidth]{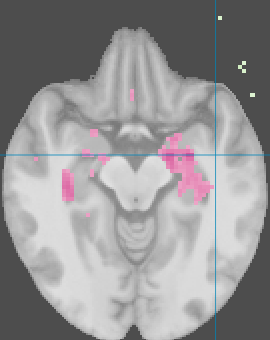} &
  \includegraphics[height=0.145\linewidth]{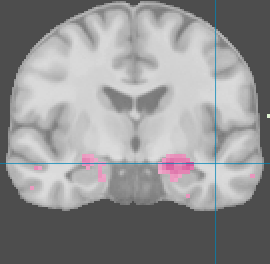} &
  \includegraphics[height=0.145\linewidth]{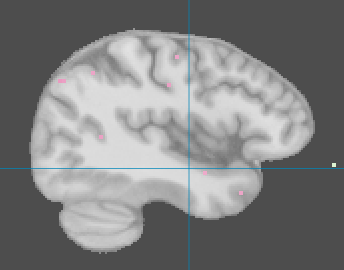} & &
  \includegraphics[height=0.145\linewidth]{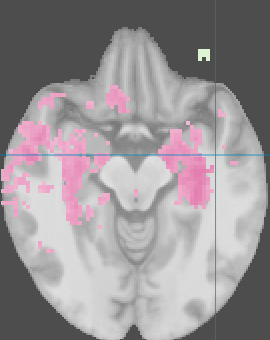} &
  \includegraphics[height=0.145\linewidth]{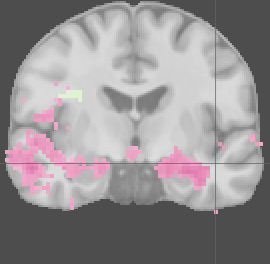} &
  \includegraphics[height=0.145\linewidth]{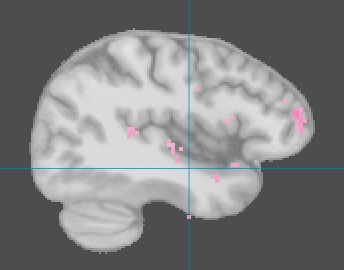} \\
  \rotatebox[origin=l]{90}{\hspace{0.5mm} \CDRm} \hspace{0.5mm} &
  \includegraphics[height=0.145\linewidth]{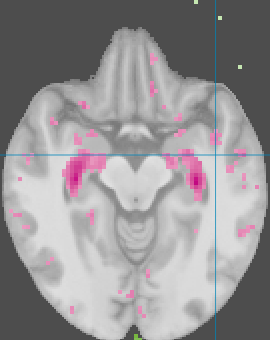} &
  \includegraphics[height=0.145\linewidth]{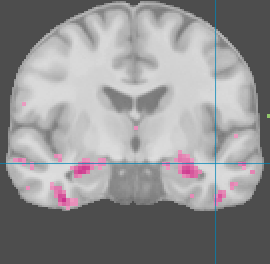} &
  \includegraphics[height=0.145\linewidth]{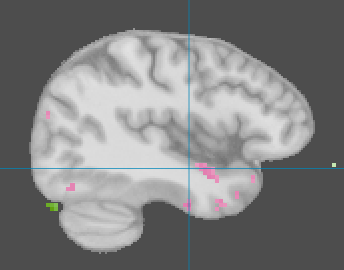} & &
  \includegraphics[height=0.145\linewidth]{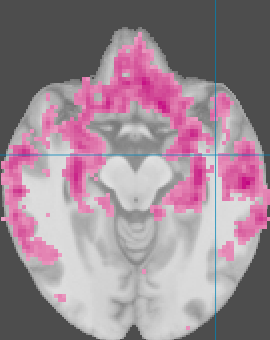} &
  \includegraphics[height=0.145\linewidth]{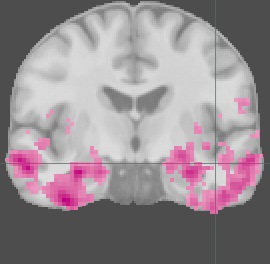} &
  \includegraphics[height=0.145\linewidth]{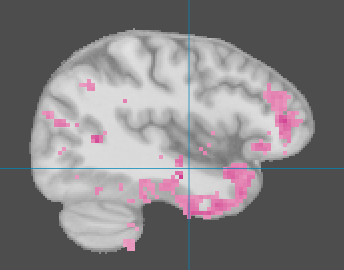} \\
  &
  \multicolumn{7}{c}{-0.62 \includegraphics[width=60mm, height=2.5mm, angle=0]{colorbar} 0.62}
\end{tabular}
\caption{\label{fig:dementia}
Axial, coronal, and sagittal slices show the \rev{inverse} correlation of gray matter tissue to
very mild (\CDRvm) and mild dementia (\CDRm).}
\end{figure}

Figure~\ref{fig:dementia} shows the results of a TBM analysis of
mild and very mild dementia with and without OTF, demonstrating 
the increased statistical power that OTF provide for spatially dispersed effects.
\TBMA~finds correlations with cognitive impairment in the temporal lobe but also
the occipital, parietal, and frontal lobe, that are not detected by a TBM
analysis without the OTF extraction step. The findings
are consistent with results from the literature~\citep{burton2004cerebral,kaye1997volume,brun1987frontal} that report gray
matter loss in the frontal, temporal, and occipital lobe. The disease
progression from CDR 0.5 to CDR 1 indicated by OTF analysis resembles
the Braak stages
from autopsies~\citep{braak1991neuropathological} that reveal a spread
from the limbic regions in the early stages to the neocortex in the later
stages. In particular the amyloid deposit stages B and C and the intra-neuronal
neurofibrillary changes stages III-IV and V-VI match the OTF analysis findings for
participants with very mild and mild dementia, respectively.

\begin{figure}[tbh]
\footnotesize
\setlength{\tabcolsep}{0.5pt}
\renewcommand{\arraystretch}{1.2}
\centering
\begin{tabular}{l ccc @{\hspace{1mm}} c @{\hspace{1mm}} ccc}
  &\multicolumn{3}{c}{TBM} & \hspace{1mm}& \multicolumn{3}{c}{\TBMA} \\[1mm]
  \rotatebox[origin=l]{90}{\hspace{0mm} \AgeH} \hspace{0.5mm} &
  \includegraphics[height=0.145\linewidth]{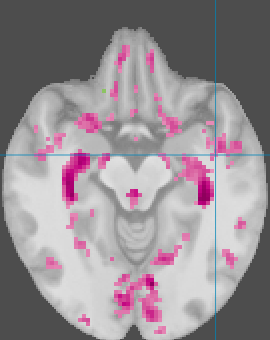} &
  \includegraphics[height=0.145\linewidth]{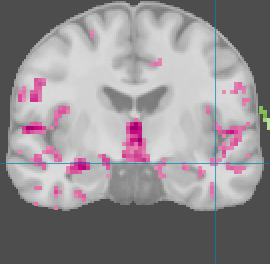} &
  \includegraphics[height=0.145\linewidth]{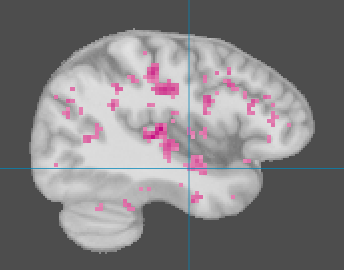} & &
  \includegraphics[height=0.145\linewidth]{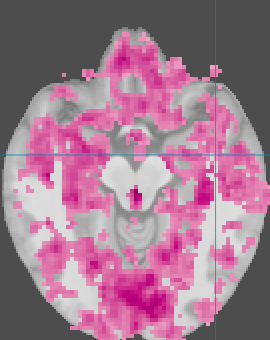} &
  \includegraphics[height=0.145\linewidth]{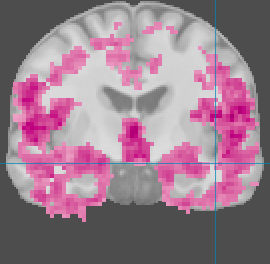} &
  \includegraphics[height=0.145\linewidth]{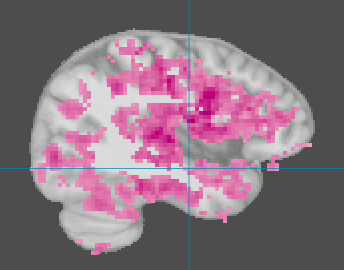} \\
  \rotatebox[origin=l]{90}{\hspace{0.5mm} \AgeD} \hspace{0.5mm} &
  \includegraphics[height=0.145\linewidth]{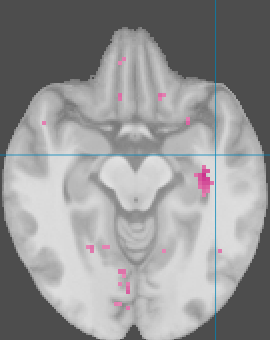} &
  \includegraphics[height=0.145\linewidth]{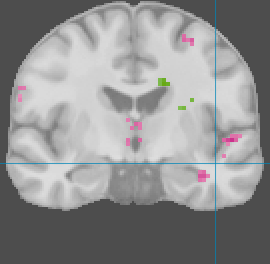} &
  \includegraphics[height=0.145\linewidth]{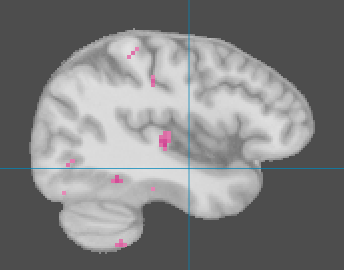} & &
  \includegraphics[height=0.145\linewidth]{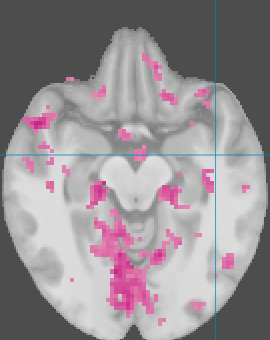} &
  \includegraphics[height=0.145\linewidth]{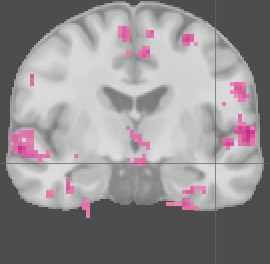} &
  \includegraphics[height=0.145\linewidth]{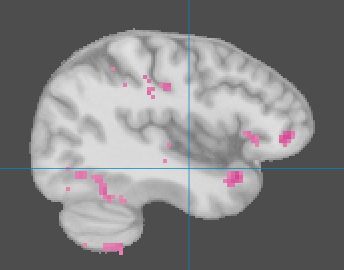} \\
  &
  \multicolumn{7}{c}{-0.75 \includegraphics[width=60mm, height=2.5mm, angle=0]{colorbar} 0.75}
\end{tabular}
\caption{\label{fig:aging}
Axial, coronal, and sagittal slices show the \rev{inverse} correlation of gray matter tissue
to \rev{age that is associated to normal aging (\AgeH{}) and to age within the dementia group (\AgeD).}}
\end{figure}

For aging, see Figure~\ref{fig:aging}, \TBMA~detects tissue loss in the
area around the postcentral gyrus. The primary somatosensory cortex, located in
this region, has been reported to correlate with age-related tissue
loss~\citep{raz1997selective}.  \TBMA~also shows loss of tissue in the
cerebellum, reported as being significantly correlated with aging in prior manual
studies~\citep{jernigan2001effects,luft1999patterns}.

The effects of normal aging and dementia differ and suggest that normal
aging is not on a
continuum with dementia-related aging,
but is rather a separate process.
Interestingly the effects of aging in the dementia group are less pronounced
than in the \rev{normal aging} group, potentially suggesting that dementia effects mask
typical aging processes. 

The \TBMA{} analysis suggests
that aging in individuals with very mild or mild dementia
progresses drastically differently from normal aging. This
difference is not clearly discernible in the \rev{\vbmaff{}} analysis. Furthermore, the effects
of aging in the dementia group (Figure~\ref{fig:aging}, bottom) are different from the effects of the
progression of cognitive impairment (Figure \ref{fig:dementia}). This could indicate that the CDR scale is not sensitive to
\rev{gray matter reductions} in certain regions and hence does not show \rev{a relationship} within the normal aging group.

\subsection{Separating Differences in Tissue Location Versus Volume}
\begin{figure}[htb]
\small
\setlength{\tabcolsep}{0.5pt}
\renewcommand{\arraystretch}{1.2}
\centering
\begin{tabular}{ cc @{\hspace{1mm}}c@{\hspace{1mm}} cc @{\hspace{1mm}}c@{\hspace{1mm}} cc}
  \multicolumn{2}{c}{TBM} & & \multicolumn{2}{c}{\TBMT} & & \multicolumn{2}{c}{\TBMA}
  \\\cline{1-2}\cline{4-5}\cline{7-8}
  \\[-3mm]
  \includegraphics[height=0.15\linewidth]{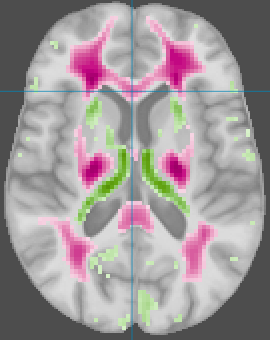} &
  \includegraphics[height=0.15\linewidth]{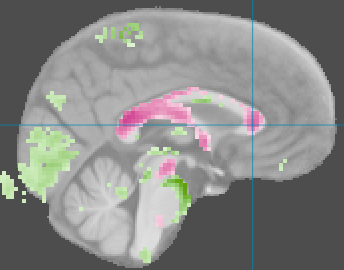} & &
  \includegraphics[height=0.15\linewidth]{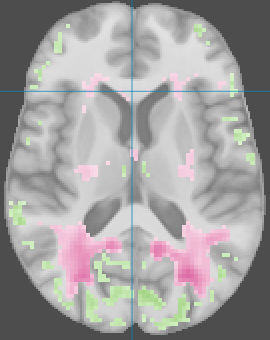} &
  \includegraphics[height=0.15\linewidth]{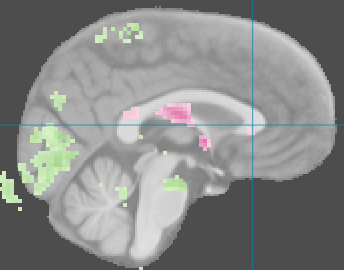} & &
  \includegraphics[height=0.15\linewidth]{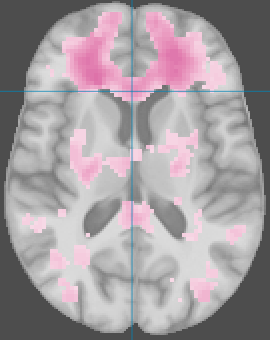} &
  \includegraphics[height=0.15\linewidth]{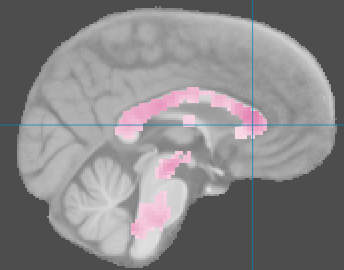} \\
  \multicolumn{8}{c}{-0.55 \includegraphics[width=60mm, height=2.5mm, angle=0]{colorbar} 0.55}
\end{tabular}
\caption{\label{fig:cor-oasis-white-1}
Axial and coronal slices displaying the results of TBM, transport cost (\TBMT),
and mass allocation (\TBMA) analysis of white matter with respect to Age.
\TBMT~indicates that some of the correlations in TBM, in particular the
correlations around the ventricles and in the posterior, are potentially
caused by differences in tissue locations between the populations and not
changes in tissue amounts. }
\end{figure}

Figure~\ref{fig:cor-oasis-white-1} shows correlation of white matter tissue to
age for TBM, \TBMT{}, and \TBMA. Correlation with transport cost (\TBMT) indicates
a consistent difference in shape or distribution of tissue location , while
correlation with mass allocation (\TBMA) indicates differences in the amount of
tissue.  The \TBMT~results indicate that some of the correlation displayed by
TBM can be explained by differences in anatomical shape and are not
necessarily associated with tissue loss: The tissue gain around the ventricles
and the occipital region in TBM can be attributed to persistent difference in
tissue location and does not correspond to tissue loss. This conclusion is
based on comparing transport cost (\TBMT) correlation to the TBM findings:
Regions where the transport cost correlations overlap with the TBM results
suggest that the TBM results are due to anatomical shape differences and/or
due to effects of the  spatial alignment. Even with a highly nonlinear warp, the
registration step is driven by tissue boundaries and leads to Jacobian
determinants that are dominant around anatomical boundaries.  For a shift in
boundaries, this results in a contraction and an expansion on opposite sides.
OTF analysis is not sensitive to shifts and mitigates this effect.

\section{Conclusion}
The work provides evidence that optimal transport features increase
the statistical power of morphometric analysis for spatially dispersed effects.
Optimal transport features capture regional and global changes while retaining
the benefits of a voxel-wise visualization and analysis of results. We
demonstrate that OTF can attribute effects to either mass transfer, due to
anatomical variations in shape or misalignment, or mass allocation, due to
volume differences. Statistical assessment demonstrates that for diffuse tissue
loss OTF analysis increases correlation strength.  The mass-balancing, introduced in
Section~\ref{sec:unbalanced}, links the OTF to VBM/TBM through a continuum that
enhances identification of effects from localized to globally dispersed.

An important step for future work is to integrate OTF to surface based
morphometric analysis. An open challenge is to consider how to apply OTF to the
analysis of functional brain images (fMRI) and other time-varying measurements.

\section*{Acknowledgments}
This work was supported by the National Institutes of Health [grant numbers R42MH118845, R01EB021391, R01HD055741, U54HD079124, R42NS086295, R44NS081792, R44CA165621, R01EB021396] and by the National Science Foundation [grant number ECCS-1711776].

%\bibliographystyle{IEEEtran}
%\bibliography{IEEEabrv,bot.bib,optimal-transport}
%\bibliographystyle{elsarticle-num}
\bibliographystyle{model2-names.bst}\biboptions{authoryear}
\bibliography{bot.bib,optimal-transport}

\begin{thebibliography}{38}
\expandafter\ifx\csname natexlab\endcsname\relax\def\natexlab#1{#1}\fi
\providecommand{\url}[1]{\texttt{#1}}
\providecommand{\href}[2]{#2}
\providecommand{\path}[1]{#1}
\providecommand{\DOIprefix}{doi:}
\providecommand{\ArXivprefix}{arXiv:}
\providecommand{\URLprefix}{URL: }
\providecommand{\Pubmedprefix}{pmid:}
\providecommand{\doi}[1]{\href{http://dx.doi.org/#1}{\path{#1}}}
\providecommand{\Pubmed}[1]{\href{pmid:#1}{\path{#1}}}
\providecommand{\bibinfo}[2]{#2}
\ifx\xfnm\relax \def\xfnm[#1]{\unskip,\space#1}\fi
%Type = Book
\bibitem[{Ahuja et~al.(1993)Ahuja, Magnanti and Orlin}]{Ahuja:1993:NFT:137406}
\bibinfo{author}{Ahuja, R.K.}, \bibinfo{author}{Magnanti, T.L.},
  \bibinfo{author}{Orlin, J.B.}, \bibinfo{year}{1993}.
\newblock \bibinfo{title}{Network Flows: Theory, Algorithms, and Applications}.
\newblock \bibinfo{publisher}{Prentice-Hall, Inc.}, \bibinfo{address}{Upper
  Saddle River, NJ, USA}.
%Type = Article
\bibitem[{Anderes et~al.(2016)Anderes, Borgwardt and
  Miller}]{anderes2016discrete}
\bibinfo{author}{Anderes, E.}, \bibinfo{author}{Borgwardt, S.},
  \bibinfo{author}{Miller, J.}, \bibinfo{year}{2016}.
\newblock \bibinfo{title}{Discrete wasserstein barycenters: Optimal transport
  for discrete data}.
\newblock \bibinfo{journal}{Mathematical Methods of Operations Research}
  \bibinfo{volume}{84}, \bibinfo{pages}{389--409}.
%Type = Article
\bibitem[{Ashburner(2007)}]{ashburner2007fast}
\bibinfo{author}{Ashburner, J.}, \bibinfo{year}{2007}.
\newblock \bibinfo{title}{A fast diffeomorphic image registration algorithm}.
\newblock \bibinfo{journal}{Neuroimage} \bibinfo{volume}{38},
  \bibinfo{pages}{95--113}.
%Type = Article
\bibitem[{Ashburner and Friston(2000)}]{ashburner2000voxel}
\bibinfo{author}{Ashburner, J.}, \bibinfo{author}{Friston, K.J.},
  \bibinfo{year}{2000}.
\newblock \bibinfo{title}{Voxel-based morphometry—the methods}.
\newblock \bibinfo{journal}{Neuroimage} \bibinfo{volume}{11},
  \bibinfo{pages}{805--821}.
%Type = Article
\bibitem[{Ashburner et~al.(1998)Ashburner, Hutton, Frackowiak, Johnsrude,
  Price, Friston et~al.}]{ashburner1998identifying}
\bibinfo{author}{Ashburner, J.}, \bibinfo{author}{Hutton, C.},
  \bibinfo{author}{Frackowiak, R.}, \bibinfo{author}{Johnsrude, I.},
  \bibinfo{author}{Price, C.}, \bibinfo{author}{Friston, K.}, et~al.,
  \bibinfo{year}{1998}.
\newblock \bibinfo{title}{Identifying global anatomical differences:
  deformation-based morphometry}.
\newblock \bibinfo{journal}{Human brain mapping} \bibinfo{volume}{6},
  \bibinfo{pages}{348--357}.
%Type = Article
\bibitem[{Avants et~al.(2008)Avants, Epstein, Grossman and
  Gee}]{avants2008symmetric}
\bibinfo{author}{Avants, B.B.}, \bibinfo{author}{Epstein, C.L.},
  \bibinfo{author}{Grossman, M.}, \bibinfo{author}{Gee, J.C.},
  \bibinfo{year}{2008}.
\newblock \bibinfo{title}{Symmetric diffeomorphic image registration with
  cross-correlation: evaluating automated labeling of elderly and
  neurodegenerative brain}.
\newblock \bibinfo{journal}{Medical image analysis} \bibinfo{volume}{12},
  \bibinfo{pages}{26--41}.
%Type = Article
\bibitem[{Benamou(2003)}]{benamou2003numerical}
\bibinfo{author}{Benamou, J.D.}, \bibinfo{year}{2003}.
\newblock \bibinfo{title}{Numerical resolution of an “unbalanced” mass
  transport problem}.
\newblock \bibinfo{journal}{ESAIM: Mathematical Modelling and Numerical
  Analysis} \bibinfo{volume}{37}, \bibinfo{pages}{851--868}.
%Type = Article
\bibitem[{Benamou and Brenier(2000)}]{benamou:nm2000}
\bibinfo{author}{Benamou, J.D.}, \bibinfo{author}{Brenier, Y.},
  \bibinfo{year}{2000}.
\newblock \bibinfo{title}{A computational fluid mechanics solution to the
  monge-kantorovich mass transfer problem}.
\newblock \bibinfo{journal}{Numerische Mathematik} \bibinfo{volume}{84},
  \bibinfo{pages}{375--393}.
%Type = Article
\bibitem[{Bookstein(2001)}]{bookstein2001voxel}
\bibinfo{author}{Bookstein, F.L.}, \bibinfo{year}{2001}.
\newblock \bibinfo{title}{“voxel-based morphometry” should not be used with
  imperfectly registered images}.
\newblock \bibinfo{journal}{Neuroimage} \bibinfo{volume}{14},
  \bibinfo{pages}{1454--1462}.
%Type = Article
\bibitem[{Braak and Braak(1991)}]{braak1991neuropathological}
\bibinfo{author}{Braak, H.}, \bibinfo{author}{Braak, E.}, \bibinfo{year}{1991}.
\newblock \bibinfo{title}{Neuropathological stageing of alzheimer-related
  changes}.
\newblock \bibinfo{journal}{Acta neuropathologica} \bibinfo{volume}{82},
  \bibinfo{pages}{239--259}.
%Type = Article
\bibitem[{Brun(1987)}]{brun1987frontal}
\bibinfo{author}{Brun, A.}, \bibinfo{year}{1987}.
\newblock \bibinfo{title}{Frontal lobe degeneration of non-alzheimer type. i.
  neuropathology}.
\newblock \bibinfo{journal}{Archives of gerontology and geriatrics}
  \bibinfo{volume}{6}, \bibinfo{pages}{193--208}.
%Type = Article
\bibitem[{Burton et~al.(2004)Burton, McKeith, Burn, Williams and
  O’Brien}]{burton2004cerebral}
\bibinfo{author}{Burton, E.J.}, \bibinfo{author}{McKeith, I.G.},
  \bibinfo{author}{Burn, D.J.}, \bibinfo{author}{Williams, E.D.},
  \bibinfo{author}{O’Brien, J.T.}, \bibinfo{year}{2004}.
\newblock \bibinfo{title}{Cerebral atrophy in parkinson’s disease with and
  without dementia: a comparison with alzheimer’s disease, dementia with lewy
  bodies and controls}.
\newblock \bibinfo{journal}{Brain} \bibinfo{volume}{127},
  \bibinfo{pages}{791--800}.
%Type = Article
\bibitem[{Chizat et~al.(2018)Chizat, Peyr{\'{e}}, Schmitzer and
  Vialard}]{Chizat_2018}
\bibinfo{author}{Chizat, L.}, \bibinfo{author}{Peyr{\'{e}}, G.},
  \bibinfo{author}{Schmitzer, B.}, \bibinfo{author}{Vialard, F.X.},
  \bibinfo{year}{2018}.
\newblock \bibinfo{title}{Unbalanced optimal transport: Dynamic and kantorovich
  formulations}.
\newblock \bibinfo{journal}{Journal of Functional Analysis}
  \bibinfo{volume}{274}, \bibinfo{pages}{3090--3123}.
%Type = Inproceedings
\bibitem[{Cuturi(2013)}]{cuturi2013sinkhorn}
\bibinfo{author}{Cuturi, M.}, \bibinfo{year}{2013}.
\newblock \bibinfo{title}{Sinkhorn distances: Lightspeed computation of optimal
  transport}, in: \bibinfo{booktitle}{Advances in neural information processing
  systems}, pp. \bibinfo{pages}{2292--2300}.
%Type = Inproceedings
\bibitem[{Cuturi and Doucet(2014)}]{cuturi2014fast}
\bibinfo{author}{Cuturi, M.}, \bibinfo{author}{Doucet, A.},
  \bibinfo{year}{2014}.
\newblock \bibinfo{title}{Fast computation of wasserstein barycenters}, in:
  \bibinfo{booktitle}{International Conference on Machine Learning}, pp.
  \bibinfo{pages}{685--693}.
%Type = Article
\bibitem[{Dale et~al.(1999)Dale, Fischl and Sereno}]{dale1999cortical}
\bibinfo{author}{Dale, A.M.}, \bibinfo{author}{Fischl, B.},
  \bibinfo{author}{Sereno, M.I.}, \bibinfo{year}{1999}.
\newblock \bibinfo{title}{Cortical surface-based analysis: I. segmentation and
  surface reconstruction}.
\newblock \bibinfo{journal}{Neuroimage} \bibinfo{volume}{9},
  \bibinfo{pages}{179--194}.
%Type = Article
\bibitem[{Davatzikos(2004)}]{davatzikos2004voxel}
\bibinfo{author}{Davatzikos, C.}, \bibinfo{year}{2004}.
\newblock \bibinfo{title}{Why voxel-based morphometric analysis should be used
  with great caution when characterizing group differences}.
\newblock \bibinfo{journal}{Neuroimage} \bibinfo{volume}{23},
  \bibinfo{pages}{17--20}.
%Type = Inproceedings
\bibitem[{Feydy et~al.(2017)Feydy, Charlier, Vialard and
  Peyr{\'e}}]{feydy2017optimal}
\bibinfo{author}{Feydy, J.}, \bibinfo{author}{Charlier, B.},
  \bibinfo{author}{Vialard, F.X.}, \bibinfo{author}{Peyr{\'e}, G.},
  \bibinfo{year}{2017}.
\newblock \bibinfo{title}{Optimal transport for diffeomorphic registration},
  in: \bibinfo{booktitle}{International Conference on Medical Image Computing
  and Computer-Assisted Intervention}, \bibinfo{organization}{Springer}. pp.
  \bibinfo{pages}{291--299}.
%Type = Book
\bibitem[{Frackowiak(2004)}]{frackowiak2004human}
\bibinfo{author}{Frackowiak, R.S.}, \bibinfo{year}{2004}.
\newblock \bibinfo{title}{Human brain function}.
\newblock \bibinfo{publisher}{Academic press}.
%Type = Article
\bibitem[{Gangbo et~al.(2019)Gangbo, Li, Osher and Puthawala}]{Gangbo_2019}
\bibinfo{author}{Gangbo, W.}, \bibinfo{author}{Li, W.}, \bibinfo{author}{Osher,
  S.}, \bibinfo{author}{Puthawala, M.}, \bibinfo{year}{2019}.
\newblock \bibinfo{title}{Unnormalized optimal transport}.
\newblock \bibinfo{journal}{Journal of Computational Physics}
  \bibinfo{volume}{399}, \bibinfo{pages}{108940}.
%Type = Article
\bibitem[{Gerber and Maggioni(2017)}]{gerber2017multiscale}
\bibinfo{author}{Gerber, S.}, \bibinfo{author}{Maggioni, M.},
  \bibinfo{year}{2017}.
\newblock \bibinfo{title}{Multiscale strategies for computing optimal
  transport}.
\newblock \bibinfo{journal}{Journal of Machine Learning Research}
  \bibinfo{volume}{18}, \bibinfo{pages}{1--32}.
%Type = Inproceedings
\bibitem[{Gerber et~al.(2018)Gerber, Niethammer, Styner and
  Aylward}]{gerber2018exploratory}
\bibinfo{author}{Gerber, S.}, \bibinfo{author}{Niethammer, M.},
  \bibinfo{author}{Styner, M.}, \bibinfo{author}{Aylward, S.},
  \bibinfo{year}{2018}.
\newblock \bibinfo{title}{Exploratory population analysis with unbalanced
  optimal transport}, in: \bibinfo{booktitle}{International Conference on
  Medical Image Computing and Computer-Assisted Intervention},
  \bibinfo{organization}{Springer}. pp. \bibinfo{pages}{464--472}.
%Type = Inproceedings
\bibitem[{Gramfort et~al.(2015)Gramfort, Peyr{\'e} and
  Cuturi}]{gramfort2015fast}
\bibinfo{author}{Gramfort, A.}, \bibinfo{author}{Peyr{\'e}, G.},
  \bibinfo{author}{Cuturi, M.}, \bibinfo{year}{2015}.
\newblock \bibinfo{title}{Fast optimal transport averaging of neuroimaging
  data}, in: \bibinfo{booktitle}{International Conference on Information
  Processing in Medical Imaging}, \bibinfo{organization}{Springer}. pp.
  \bibinfo{pages}{261--272}.
%Type = Phdthesis
\bibitem[{Guittet(2002)}]{guittet2002extended}
\bibinfo{author}{Guittet, K.}, \bibinfo{year}{2002}.
\newblock \bibinfo{title}{Extended Kantorovich norms: a tool for optimization}.
\newblock Ph.D. thesis. INRIA.
%Type = Article
\bibitem[{Hua et~al.(2008)Hua, Leow, Parikshak, Lee, Chiang, Toga, Jack~Jr,
  Weiner, Thompson, Initiative et~al.}]{hua2008tensor}
\bibinfo{author}{Hua, X.}, \bibinfo{author}{Leow, A.D.},
  \bibinfo{author}{Parikshak, N.}, \bibinfo{author}{Lee, S.},
  \bibinfo{author}{Chiang, M.C.}, \bibinfo{author}{Toga, A.W.},
  \bibinfo{author}{Jack~Jr, C.R.}, \bibinfo{author}{Weiner, M.W.},
  \bibinfo{author}{Thompson, P.M.}, \bibinfo{author}{Initiative, A.D.N.},
  et~al., \bibinfo{year}{2008}.
\newblock \bibinfo{title}{Tensor-based morphometry as a neuroimaging biomarker
  for alzheimer's disease: an mri study of 676 ad, mci, and normal subjects}.
\newblock \bibinfo{journal}{Neuroimage} \bibinfo{volume}{43},
  \bibinfo{pages}{458--469}.
%Type = Article
\bibitem[{Jernigan et~al.(2001)Jernigan, Archibald, Fennema-Notestine, Gamst,
  Stout, Bonner and Hesselink}]{jernigan2001effects}
\bibinfo{author}{Jernigan, T.L.}, \bibinfo{author}{Archibald, S.L.},
  \bibinfo{author}{Fennema-Notestine, C.}, \bibinfo{author}{Gamst, A.C.},
  \bibinfo{author}{Stout, J.C.}, \bibinfo{author}{Bonner, J.},
  \bibinfo{author}{Hesselink, J.R.}, \bibinfo{year}{2001}.
\newblock \bibinfo{title}{Effects of age on tissues and regions of the cerebrum
  and cerebellum}.
\newblock \bibinfo{journal}{Neurobiology of aging} \bibinfo{volume}{22},
  \bibinfo{pages}{581--594}.
%Type = Article
\bibitem[{Kaye et~al.(1997)Kaye, Swihart, Howieson, Dame, Moore, Karnos,
  Camicioli, Ball, Oken and Sexton}]{kaye1997volume}
\bibinfo{author}{Kaye, J.A.}, \bibinfo{author}{Swihart, T.},
  \bibinfo{author}{Howieson, D.}, \bibinfo{author}{Dame, A.},
  \bibinfo{author}{Moore, M.}, \bibinfo{author}{Karnos, T.},
  \bibinfo{author}{Camicioli, R.}, \bibinfo{author}{Ball, M.},
  \bibinfo{author}{Oken, B.}, \bibinfo{author}{Sexton, G.},
  \bibinfo{year}{1997}.
\newblock \bibinfo{title}{Volume loss of the hippocampus and temporal lobe in
  healthy elderly persons destined to develop dementia}.
\newblock \bibinfo{journal}{Neurology} \bibinfo{volume}{48},
  \bibinfo{pages}{1297--1304}.
%Type = Article
\bibitem[{Kundu et~al.(2018)Kundu, Kolouri, Erickson, Kramer, McAuley and
  Rohde}]{kundu2018discovery}
\bibinfo{author}{Kundu, S.}, \bibinfo{author}{Kolouri, S.},
  \bibinfo{author}{Erickson, K.I.}, \bibinfo{author}{Kramer, A.F.},
  \bibinfo{author}{McAuley, E.}, \bibinfo{author}{Rohde, G.K.},
  \bibinfo{year}{2018}.
\newblock \bibinfo{title}{Discovery and visualization of structural biomarkers
  from mri using transport-based morphometry}.
\newblock \bibinfo{journal}{NeuroImage} \bibinfo{volume}{167},
  \bibinfo{pages}{256--275}.
%Type = Article
\bibitem[{Lee et~al.(2021)Lee, Lai, Li and Osher}]{Lee_2021}
\bibinfo{author}{Lee, W.}, \bibinfo{author}{Lai, R.}, \bibinfo{author}{Li, W.},
  \bibinfo{author}{Osher, S.}, \bibinfo{year}{2021}.
\newblock \bibinfo{title}{Generalized unnormalized optimal transport and its
  fast algorithms}.
\newblock \bibinfo{journal}{Journal of Computational Physics}
  \bibinfo{volume}{436}, \bibinfo{pages}{110041}.
%Type = Article
\bibitem[{Liero et~al.(2017)Liero, Mielke and Savar{\'{e}}}]{Liero_2017}
\bibinfo{author}{Liero, M.}, \bibinfo{author}{Mielke, A.},
  \bibinfo{author}{Savar{\'{e}}, G.}, \bibinfo{year}{2017}.
\newblock \bibinfo{title}{Optimal entropy-transport problems and a new
  hellinger{\textendash}kantorovich distance between positive measures}.
\newblock \bibinfo{journal}{Inventiones mathematicae} \bibinfo{volume}{211},
  \bibinfo{pages}{969--1117}.
%Type = Article
\bibitem[{Luft et~al.(1999)Luft, Skalej, Schulz, Welte, Kolb, B{\"u}rk,
  Klockgether and Voigt}]{luft1999patterns}
\bibinfo{author}{Luft, A.R.}, \bibinfo{author}{Skalej, M.},
  \bibinfo{author}{Schulz, J.B.}, \bibinfo{author}{Welte, D.},
  \bibinfo{author}{Kolb, R.}, \bibinfo{author}{B{\"u}rk, K.},
  \bibinfo{author}{Klockgether, T.}, \bibinfo{author}{Voigt, K.},
  \bibinfo{year}{1999}.
\newblock \bibinfo{title}{Patterns of age-related shrinkage in cerebellum and
  brainstem observed in vivo using three-dimensional mri volumetry}.
\newblock \bibinfo{journal}{Cerebral Cortex} \bibinfo{volume}{9},
  \bibinfo{pages}{712--721}.
%Type = Article
\bibitem[{Marcus et~al.(2010)Marcus, Fotenos, Csernansky, Morris and
  Buckner}]{marcus2010open}
\bibinfo{author}{Marcus, D.S.}, \bibinfo{author}{Fotenos, A.F.},
  \bibinfo{author}{Csernansky, J.G.}, \bibinfo{author}{Morris, J.C.},
  \bibinfo{author}{Buckner, R.L.}, \bibinfo{year}{2010}.
\newblock \bibinfo{title}{Open access series of imaging studies: longitudinal
  mri data in nondemented and demented older adults}.
\newblock \bibinfo{journal}{Journal of cognitive neuroscience}
  \bibinfo{volume}{22}, \bibinfo{pages}{2677--2684}.
%Type = Incollection
\bibitem[{M{\'{e}}rigot and Thibert(2021)}]{Merigot_2021}
\bibinfo{author}{M{\'{e}}rigot, Q.}, \bibinfo{author}{Thibert, B.},
  \bibinfo{year}{2021}.
\newblock \bibinfo{title}{Optimal transport: discretization and algorithms},
  in: \bibinfo{booktitle}{Geometric Partial Differential Equations - Part
  {II}}. \bibinfo{publisher}{Elsevier}, pp. \bibinfo{pages}{133--212}.
%Type = Article
\bibitem[{Prakash et~al.(2022)Prakash, Balakrishna and
  Thenepalle}]{Prakash_2022}
\bibinfo{author}{Prakash, A.}, \bibinfo{author}{Balakrishna, U.},
  \bibinfo{author}{Thenepalle, J.K.}, \bibinfo{year}{2022}.
\newblock \bibinfo{title}{An exact algorithm for constrained k-cardinality
  unbalanced assignment problem}.
\newblock \bibinfo{journal}{International Journal of Industrial Engineering
  Computations} \bibinfo{volume}{13}, \bibinfo{pages}{267--276}.
%Type = Article
\bibitem[{Raz et~al.(1997)Raz, Gunning, Head, Dupuis, McQuain, Briggs, Loken,
  Thornton and Acker}]{raz1997selective}
\bibinfo{author}{Raz, N.}, \bibinfo{author}{Gunning, F.M.},
  \bibinfo{author}{Head, D.}, \bibinfo{author}{Dupuis, J.H.},
  \bibinfo{author}{McQuain, J.}, \bibinfo{author}{Briggs, S.D.},
  \bibinfo{author}{Loken, W.J.}, \bibinfo{author}{Thornton, A.E.},
  \bibinfo{author}{Acker, J.D.}, \bibinfo{year}{1997}.
\newblock \bibinfo{title}{Selective aging of the human cerebral cortex observed
  in vivo: differential vulnerability of the prefrontal gray matter.}
\newblock \bibinfo{journal}{Cerebral cortex (New York, NY: 1991)}
  \bibinfo{volume}{7}, \bibinfo{pages}{268--282}.
%Type = Article
\bibitem[{Rohlfing et~al.(2010)Rohlfing, Zahr, Sullivan and
  Pfefferbaum}]{rohlfing2010sri24}
\bibinfo{author}{Rohlfing, T.}, \bibinfo{author}{Zahr, N.M.},
  \bibinfo{author}{Sullivan, E.V.}, \bibinfo{author}{Pfefferbaum, A.},
  \bibinfo{year}{2010}.
\newblock \bibinfo{title}{The sri24 multichannel atlas of normal adult human
  brain structure}.
\newblock \bibinfo{journal}{Human brain mapping} \bibinfo{volume}{31},
  \bibinfo{pages}{798--819}.
%Type = Article
\bibitem[{Scarpazza and De~Simone(2016)}]{scarpazza2016voxel}
\bibinfo{author}{Scarpazza, C.}, \bibinfo{author}{De~Simone, M.S.},
  \bibinfo{year}{2016}.
\newblock \bibinfo{title}{Voxel-based morphometry: current perspectives}.
\newblock \bibinfo{journal}{Neuroscience and Neuroeconomics}
  \bibinfo{volume}{5}, \bibinfo{pages}{19--35}.
%Type = Book
\bibitem[{Talairach and Tournoux(1988)}]{talaraich:book88}
\bibinfo{author}{Talairach, J.}, \bibinfo{author}{Tournoux, P.},
  \bibinfo{year}{1988}.
\newblock \bibinfo{title}{Co-planar stereotaxic atlas of the human brain.
  3-Dimensional proportional system: an approach to cerebral imaging}.
\newblock \bibinfo{publisher}{Thieme}, \bibinfo{address}{New York}.

\end{thebibliography}

\end{document}